\documentclass[acmlarge,nonacm]{acmart}

\AtBeginDocument{%
  }

\setcopyright{cc}
\setcctype{by}
\acmJournal{TCPS}
\acmYear{2025} \acmVolume{1} \acmNumber{1} \acmArticle{1}
\acmMonth{1}\acmDOI{10.1145/3773282}

\acmISBN{978-1-4503-XXXX-X/18/06}

\usepackage[utf8]{inputenc} % allow utf-8 input
\usepackage[T1]{fontenc}    % use 8-bit T1 fonts
\usepackage{booktabs}       % professional-quality tables
\usepackage{amsfonts}       % blackboard math symbols
\usepackage{nicefrac}       % compact symbols for 1/2, etc.
\usepackage{microtype}      % microtypography
\usepackage{xcolor}         % colors

\usepackage{siunitx}
\DeclareSIUnit{\belmilliwatt}{Bm}
\DeclareSIUnit{\dBm}{\deci\belmilliwatt}

\usepackage{graphicx}
\usepackage{booktabs}
\newcommand{\ra}[1]{\renewcommand{\arraystretch}{#1}}

\usepackage{makecell}

\usepackage{tabularx,colortbl}
\usepackage{pifont}

\usepackage{algorithm} 
\usepackage{algpseudocode}

\usepackage{pgfplots}
\usepackage{tikz}

\usepackage{caption}
\usepackage{subcaption}

\usepackage{amsmath}

\usepackage{wrapfig}

\usepackage{multirow}

\usepackage{hhline}

\usepackage{enumitem}

\usepackage[normalem]{ulem}

\usetikzlibrary{shapes}
\usetikzlibrary{positioning}
\usetikzlibrary{arrows,decorations.markings}
\usetikzlibrary{calc,spy,shapes}
\usetikzlibrary{shapes}
\usepgfplotslibrary{fillbetween}

\usetikzlibrary{pgfplots.groupplots}
\usetikzlibrary{patterns}

\newcommand{\fakepar}[1]{\vspace{0mm}\noindent\textbf{#1.}}

\newcommand{\capt}[1]{\textit{#1}}

\newcommand{\plotfontsize}{8pt}
\newcommand{\ylabelyshift}{-5mm}
\newcommand{\xlabelyshift}{2.5mm}

\newcommand{\challengelowpower}[0]{\hyperref[cha:lowpower]{\textbf{C1}}}
\newcommand{\challengebandwidth}[0]{\hyperref[cha:bandwidth]{\textbf{C2}}}

\newcommand{\propertyefficiency}[0]{\hyperref[pro:efficiency]{\textbf{P1}}}
\newcommand{\propertyparallel}[0]{\hyperref[pro:parallel]{\textbf{P2}}}

\newcommand{\requirementalltoall}[0]{\hyperref[req:alltoall]{\textbf{R1}}}
\newcommand{\requirementreliable}[0]{\hyperref[req:reliable]{\textbf{R2}}}
\newcommand{\requirementsync}[0]{\hyperref[req:sync]{\textbf{R3}}}

\newcommand{\realnmbr}{\mathbb{R}}

\newcommand{\plotheightresources}{0.6\linewidth}
\newcommand{\plotwidthresources}{0.7\linewidth}

\newcommand{\opindent}{\phantom{=}}

\begin{document}

\title{\textsc{RockNet}: Distributed Learning on Ultra-Low-Power Devices}

\author{Alexander Gräfe}
\email{alexander.graefe@dsme.rwth-aachen.de}
\affiliation{%
  \institution{Institute for Data Science in Mechanical Engineering, RWTH Aachen University}
  \city{Aachen}
  \country{Germany}
}

\author{Fabian Mager}
\affiliation{%
  \institution{Networked Embedded Systems Lab, TU Darmstadt}
  \city{Darmstadt}
  \country{Germany}
}

\author{Marco Zimmerling}
\affiliation{%
  \institution{Networked Embedded Systems Lab, TU Darmstadt}
  \city{Darmstadt}
  \country{Germany}
}

\author{Sebastian Trimpe}
\affiliation{%
\institution{Institute for Data Science in Mechanical Engineering, RWTH Aachen University}
\city{RWTH Aachen University}
\country{Germany}
}

\renewcommand{\shortauthors}{Gräfe et al.}

\begin{abstract}
  As Machine Learning (ML) becomes integral to Cyber-Physical Systems (CPS), there is growing interest in shifting training from traditional cloud-based to on-device processing (TinyML), for example, due to privacy and latency concerns.
  However, CPS often comprise ultra-low-power microcontrollers, whose limited compute resources make training challenging.
  This paper presents \textsc{RockNet}, a new TinyML method tailored for ultra-low-power hardware that achieves state-of-the-art accuracy in timeseries classification, such as fault or malware detection, without requiring offline pretraining. 
  By leveraging that CPS consist of multiple devices, we design a distributed learning method that integrates ML and wireless communication.
  \textsc{RockNet} leverages all devices for distributed training of specialized compute efficient classifiers that need minimal communication overhead for parallelization.
  Combined with tailored and efficient wireless multi-hop communication protocols, our approach overcomes the communication bottleneck that often occurs in distributed learning.
  Hardware experiments on a testbed with 20 ultra-low-power devices demonstrate \textsc{RockNet}'s effectiveness. 
  It successfully learns timeseries classification tasks from scratch, surpassing the accuracy of the latest approach for neural network microcontroller training by up to 2x.
  \textsc{RockNet}'s distributed ML architecture reduces memory, latency and energy consumption per device by up to 90 \% when scaling from one central device to 20 devices.
  Our results show that a tight integration of distributed ML, distributed computing, and communication enables, for the first time, training on ultra-low-power hardware with state-of-the-art accuracy.
\end{abstract}





%    \received{14 November 2024}

    \maketitle

\section{Introduction}  
Machine Learning (ML) is a key component of Cyber-Physical Systems (CPS)~\cite{mohammadi2018deep,fei2019cps,liang2019machine,javaid2022artificial,atat2018big}, enabling them to adapt to new situations, extract valuable insights from their data, and increase their overall efficiency.
Take smart devices such as cordless power tools in manufacturing~\cite{baumann2020wireless} as an example. 
To minimize downtimes, the tools should be able to predict their wear and tear for predictive maintenance~\cite{saxena2020iot}.
ML enables this by collecting data from all tools in the factory and extracting accurate wear models.
Another example are wearable devices equipped with multiple inertial sensors~\cite{teufl2018towards,beuchert2020overcoming} that analyze gaits to evaluate orthoses, prosthetics,
surgical procedures~\cite{chen2016toward} or to detect diseases~\cite{trabassi2022machine}.
Here, ML creates models that accurately analyze gait patterns using data from multiple users~\cite{marimon2024kinematic,trabassi2022machine}.
Other applications where ML enhances CPS include forest observation or wildfire detection, production, healthcare, smart cities, homes and power grids~\cite{mohammadi2018deep,fei2019cps,liang2019machine,olowononi2020resilient, atat2018big, werner2024ecosense, tuncel2023self}.

In the context of ML in CPS, it is crucial to differentiate between inference--the execution of a trained model--and online learning, which trains the model at runtime. Unlike inference, learning enables systems to adjust to changing operating conditions~\cite{llisterri2022device,ren2021tinyol}.
Thus, learning provides a crucial foundation for flexible and self-adapting CPS and is a focus of this work.

There exist two ways to execute learning: \emph{inside} a CPS on its devices or \emph{outside} on a cloud server, see, e.g., the surveys~\cite{atat2018big,fei2019cps,ren2023survey,dutta2021tinyml}.
The strength of a cloud server lies in its significantly greater compute power compared to the CPS devices~\cite{fei2019cps,ren2023survey,dutta2021tinyml}. On the other hand, cloud processing comes with increased latency due to the communication between the CPS and the cloud~\cite{mohammadi2018deep,fei2019cps,ren2023survey,dutta2021tinyml}, overloads existing Internet communication infrastructure due to the rapidly increasing device numbers~\cite{ciscoedge,le2024applications}, and raises privacy concerns from transmitting data outside the CPS~\cite{ren2023survey,liang2019machine,dutta2021tinyml,le2024applications}. 

Given the relevance of these issues across various applications, there is an increasing interest in implementing learning \emph{inside} CPS~\cite{mohammadi2018deep,dutta2021tinyml}.
Yet, many of the mentioned applications share the common feature of utilizing low-cost, ultra-low-power embedded devices, such as those found in smart sensors.
These have severely constrained compute resources that are challenging for ML~\cite{lin2022device,dutta2021tinyml,shafique2021tinyml}.
The research field aiming to address learning under such limited resources is known as TinyML.
However, TinyML often sacrifices learning accuracy for resource efficiency~\cite{dutta2021tinyml}, leading to suboptimal performance when running on ultra-low-power hardware.
Additionally, many works on TinyML focus on the efficient execution of \emph{already trained} models, whereas
we investigate \emph{actual learning}, a less explored TinyML branch~\cite{llisterri2022device}.
To summarize, we ask: \emph{Can we design a TinyML method for CPS that achieves state-of-the-art accuracy while learning on ultra-low-power hardware?}

\subsection{Problem Setting}
\label{sec:problemdefinition}

Abstracting from the aforementioned application examples, we consider scenarios consisting of $N$ ultra-low-power devices
together forming one CPS, like multiple tools in a smart factory or multiple smart devices in a smart home.
Each device $i$ holds local data $\mathcal{D}_i$ that it has collected.
Each entry in $\mathcal{D}_i$ consists of a datapoint $x$ with its class label~$c$. 
The data contains sensitive information and should not be shared outside the CPS, making cloud processing not possible. 
However, contrasting federated learning scenarios, it is allowed to share data between devices as they lie in the same CPS.

For instance, in the aforementioned factory setting, each power tool collects and stores timeseries sensor data, such as accelerometer readings~\cite{giordano2023optimizing}.
While the data from the tools should not leave the factory to not reveal details about production, it is safe to share between the tools.
The same holds for a smart home. The data should not leave the home to not reveal information about its residents, but can be shared among devices inside the smart home.

Goal of the devices is to predict the class label of a given datapoint $x$.
For this, the devices train a classifier $\hat{c}=h_W(x)$ on this data, where $W$ are its trainable parameters and $\hat{c}$ is the predicted class.
The goal of the learning process is to determine parameters $W$ that minimize:
\begin{equation}
  \label{eq:loss}
  L(W) = \sum_{i=1}^{N}\sum_{(x,c)\in\mathcal{D}_i}\ell(h_{W}(x),c),
\end{equation}
where $\ell$ is a suitable loss function, e.g., the cross-entropy loss~\cite{bishop2024deeplearning}.
The loss $L$ incorporates the data of all devices. 
This is crucial for the model to generalize effectively across diverse data distributions, e.g., to handle local data shifts.

As the classifier must perform well on the combined data, the devices must communicate.
For this, they use wireless communication to increase flexibility and cost efficiency compared to wired communication~\cite{baumann2020wireless}.
In many CPS applications, like large factories or underground mining, the devices have to cover large areas~\cite{baumann2020wireless}. 
Thus, not every device may have a direct link to each other device.
Devices with no direct link can communicate using devices in-between as relays (multi-hop communication).
When the devices are mobile, like wearables or robot swarms, the network topology changes continuously.

Such a setting poses two major challenges:
\setlist[itemize,1]{leftmargin=2em}
\begin{itemize}
  \item[\challengelowpower] \label{cha:lowpower} \textbf{ML on Ultra-low-power Devices.} Ultra-low-power devices are limited in energy, compute power and memory. This significantly hinders the training process, as limited compute power and energy cause delays. Furthermore, the available memory may be insufficient to store and train the entire classifier, which is directly linked to accuracy, as usually larger models lead to higher accuracies. 
  \item[\challengebandwidth] \label{cha:bandwidth} \textbf{ML with Low-power Communication.} 
  To accommodate limited device energy, we must rely on low-power communication. However, this choice inherently restricts bandwidth and slows data exchange, thus impeding the learning process—a challenge that is further exacerbated by multi-hop communication.
\end{itemize}

\fakepar{Problem Statement} In summary, our objective is to develop a classifier $h_W$ and a learning algorithm that yields parameters $W$ minimizing loss~(\ref{eq:loss}) from distributed data $\mathcal{D}_i$.
The classifier and learning algorithm shall be implemented inside the CPS on the devices communicating via a wireless multi-hop network while effectively addressing challenges~\challengelowpower{} and~\challengebandwidth{}.

\subsection{Contributions}
\label{sec:introduction:contributions}

\begin{figure}[t]
  \centering
  \begin{subfigure}[t]{0.3\linewidth}
  \centering
  \includegraphics[width=0.99\linewidth]{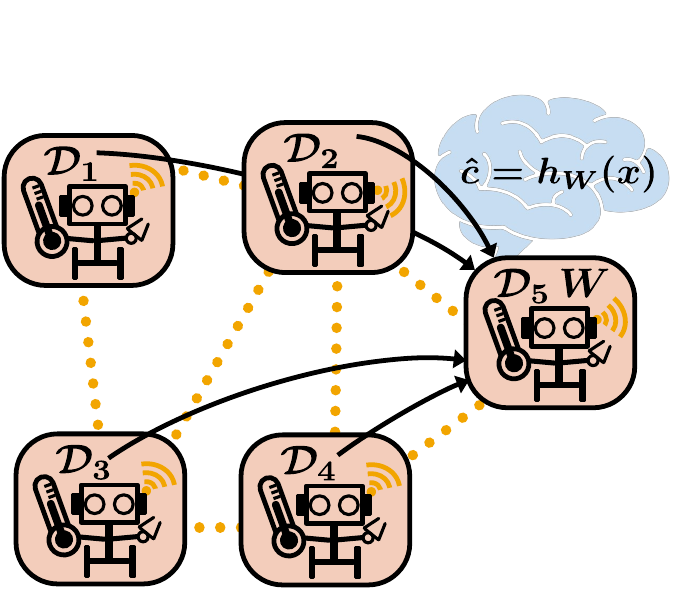}
  \vspace{-0.0cm}
  \caption{Central TinyML}
  \end{subfigure}
  \hspace{1em}
  \begin{subfigure}[t]{0.3\linewidth}
    \centering
    \includegraphics[width=0.99\linewidth]{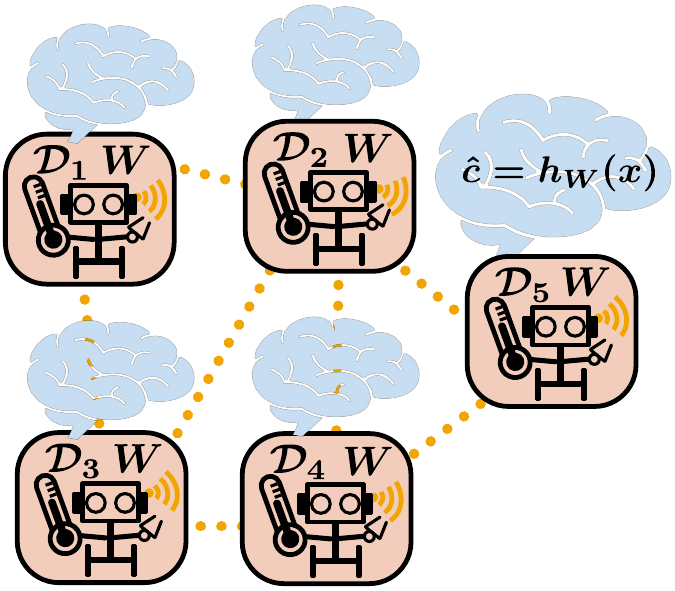}
    \vspace{-0.0cm}
    \caption{Federated TinyML}
    \end{subfigure}
    \hspace{1em}
    \begin{subfigure}[t]{0.3\linewidth}
      \centering
      \includegraphics[width=0.99\linewidth]{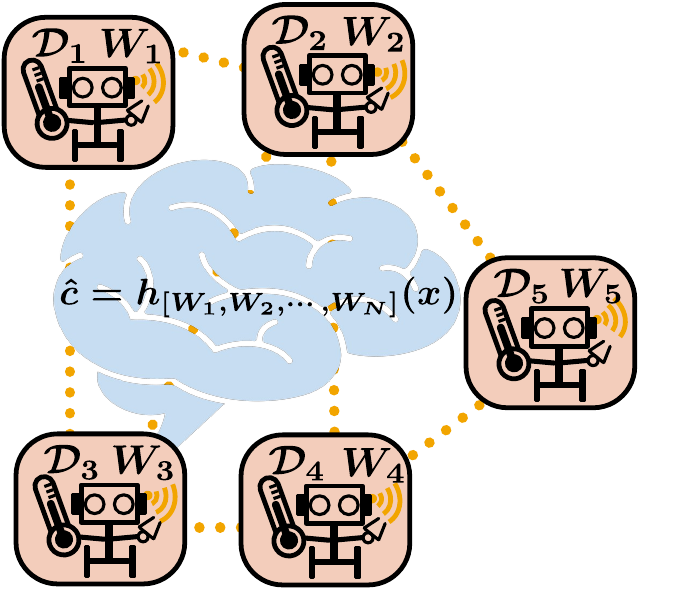}
      \vspace{-0.0cm}
      \caption{Split TinyML}
      \end{subfigure}
  \vspace{-0.4cm}
  \caption{Central versus Federated versus Split TinyML. \capt{Central TinyML: All data is sent to one device that runs the ML algorithm. Federated TinyML: Each device trains a model on its local data; these models are repeatedly shared and merged. Split TinyML: Devices collaboratively run the ML algorithm, each holding portions of the parameters $W$.
   }}
  \label{fig:setup}
  \vspace{-0.3cm}
\end{figure}

For this problem, three fundamental approaches are conceivable (Fig.~\ref{fig:setup}):
\begin{enumerate}
  \item \textbf{Central TinyML.} All devices transmit their data to one central device that executes the entire learning.
  \item \textbf{Federated TinyML.} Every device trains the entire model locally on its data. The devices repeatedly share and merge their models~\cite{hard2018federated, gao2020end, kopparapu2021tinyfedtl, kopparapu2022tinyfedtl,llisterri2022device,wulfert2023tinyfl,gimenez2023embedded}.
  \item \textbf{Split TinyML.} The devices split model execution and training among themselves. Every device $i$ holds parts $W_i$ of the classifier's parameters $W$~\cite{koda2019one, poirot2019split,vepakomma2018split, thapa2022splitfed, gao2020end}.
\end{enumerate}

Central TinyML's main limitation is its reliance on single-device execution, which inherently restricts resource availability. Consequently, models are confined in size and accuracy, and the training process is slowed due to these resource constraints.

To overcome these single-device limitations, distributed learning techniques such as federated learning and split learning pool resources of multiple devices within the CPS. Distributed learning has been highlighted as a key focus area for future research, as emphasized in ``Control for Societal-scale Challenges: Road Map 2030''~\cite{annaswamy2024control}.

While the terms "federated learning" and "split learning" are sometimes used interchangeably, we adopt the definitions provided by Gao et al.~\cite{gao2020end} and Thapa et al.~\cite{thapa2022splitfed}. 
\textbf{Federated learning} employs data parallelism: each device trains the entire model on its local data and shares the resulting model with other devices.
In contrast, \textbf{split learning} utilizes model parallelism, where each device trains different segments of the model and shares intermediate results.

Federated learning accelerates model training by enabling devices to train in parallel~\cite{wang2024achieving}.
However, each device must possess sufficient resources—particularly memory—to handle local training of the entire model~\cite{thapa2022splitfed}. 
This requirement limits federated TinyML to smaller models, often resulting in reduced accuracy.

By distributing compute and memory demands across multiple devices, split learning overcomes these limitations, enabling the training of larger, more accurate models. However, despite these inherent advantages of split TinyML and extensive research on central and federated TinyML in recent years~\cite{cai2020tinytl, de2022device, de2023mu, keshavarz2020sefr, disabato2020incremental, lin2022device, ren2021tinyol, profentzas2022minilearn, sudharsan2021ml, wulfert2024aifes, hard2018federated, gao2020end, gimenez2023embedded}, achieving split TinyML on ultra-low-power hardware remains an open challenge.

Motivated by this, we present \textsc{RockNet}, the first split TinyML method that achieves state-of-the-art accuracy while training solely and from scratch on multiple ultra-low-power devices \emph{without pretraining}.
\textsc{RockNet} is a method for timeseries classification tasks common in CPS~\cite{fei2019cps,mohammadi2018deep,liang2019machine}, e.g., anomaly detection, malware detection, fault detection and activity classification~\cite{mohammadi2018deep,giordano2023optimizing,liang2019machine}. 

Core idea behind \textsc{RockNet} is its integration of ML and communication concepts.
Specifically, \textsc{RockNet} builds on two main ideas:
\begin{enumerate}
  \item \textbf{ROCKET Classifiers Enable Efficient Split Learning:}  
  \textsc{RockNet} builds on \textit{Random Convolutional Kernel Transform} (ROCKET) classifiers~\cite{dempster2020rocket}, which achieve state-of-the-art accuracy in timeseries classification. 
  Although these classifiers are known for their resource efficiency, they remain too demanding for a single ultra-low-power device.
  However, by employing an appropriate parallelization strategy, they can be efficiently distributed among multiple devices reducing the per-device resource consumption to a level suitable for typical ultra-low-power hardware (\challengelowpower{}).
  Importantly, this parallelization strategy incurs minimal communication overhead, allowing us to meet communication constraints (\challengebandwidth{}).  
  \item \textbf{The Mixer Protocol Supports the Parallelization of ROCKET Classifiers.}  
  The second key ingredient of \textsc{RockNet} is the communication protocol Mixer~\cite{Mixer}.
  This protocol is capable of bridging the gap between the parallelization strategy and multi-hop communication. 
  It efficiently implements all-to-all communication, ensures precise network-wide synchronization, and guarantees reliable message exchange during training, all essential for our parallelization strategy. 
  Furthermore, its optimal scalability with the number of devices enables effective utilization of communication resources (\challengebandwidth{}).
\end{enumerate}

\textsc{RockNet} combines these two methods, ROCKET and Mixer, into a cohesive framework.
In a hardware experiment with up to 20 devices (nRF52840, ARM Cortex M4, \SI{64}{\mega\hertz}, \SI{256}{\kilo\byte} RAM) communicating via the physical layer of Bluetooth Low Energy, we demonstrate that \textsc{RockNet} enables ultra-low-power split TinyML for the first time.
\textsc{RockNet}'s accuracy significantly surpasses the accuracy of AIfES, a recent central TinyML approach for neural network (NN) training~\cite{wulfert2024aifes} and significantly reduces memory, latency, and energy consumption when scaling from few to many devices.

In summary, \textsc{RockNet} addresses the challenges associated with high-accuracy learning in ultra-low-power networks. 
It achieves training of high-accuracy classifiers on ultra-low-power devices and advances the state-of-the-art in split TinyML.
It makes the following contributions:
\begin{enumerate}
  \item \textsc{RockNet} is the first TinyML method to achieve state-of-the-art accuracy in timeseries classification while training on ultra-low-power hardware. 
  We enable this by integrating communication and ML research.
  \item \textsc{RockNet} is the first split learning method for decentralized ultra-low-power hardware communicating via a low-power wireless multi-hop network.
  It suits the conditions in many CPS scenarios as it does not rely on specific network topologies and supports moving devices.
  \item In experiments on real wireless hardware, we demonstrate the first learning of a non-linear timeseries classifier with state-of-the-art accuracy on ultra-low-power devices without offline pretraining. \textsc{RockNet}'s accuracy is significantly larger than central NN training. When scaling from one to 20 devices, \textsc{RockNet} reduces memory per device by up to \SI{93}{\percent}, latency by up to \SI{89}{\percent} and energy per device by up to \SI{86}{\percent}.
\end{enumerate}

\section{Related Work}

\begin{table}[t]
  \caption{Comparison of \textsc{RockNet} to related work on TinyML.}
  \label{tab:relatedwork}
  \vspace{-0.4cm}
	\centering
  \fontsize{8pt}{8pt}\selectfont
	\newcommand{\ch}{10pt}
    \ra{1.2}
	\setlength{\tabcolsep}{0.4em}
    \begin{tabular}{@{}cc||c|c|c@{}}
      \multicolumn{5}{c}{\makecell{\hspace{3cm}\textbf{	$\longleftarrow$ Methodology$\longrightarrow$}}\vspace{0.1cm}}\\
      \multirow{3}{*}{\rotatebox[origin=c]{90}{\hspace{-3cm}\textbf{$\leftarrow$Hardware$\rightarrow$}}}&&\multirow{2}{*}{\makecell{\rule{0pt}{0pt}\\\textbf{Central}\\\textbf{TinyML}}}&\multicolumn{2}{c}{\makecell{\textbf{Distributed}\\\textbf{TinyML}}}\\
      &&&\makecell{Federated\\TinyML}&\makecell{Split\\TinyML}\\
      \hhline{~====}
    &\makecell{\rule{0pt}{\ch}\textbf{Powerful}\\\textbf{Embedded}\\\textbf{Hardware}
    }
  &\cite{matsutani2022device, sudharsan2021imbal, ravaglia2021tinyml}&\makecell{\cite{hard2018federated, gao2020end,wulfert2023adaptive}\\\cite{mathur2021device,jiang2022model, sudharsan2022elastiquant}}&\cite{gao2020end,gao2021evaluation}\\
  \cline{2-5}
	&\makecell{\textbf{Ultra-}\\\textbf{low-power}\\\textbf{Hardware}}&\makecell{\rule{0pt}{\ch}\cite{cai2020tinytl, de2022device, de2023mu}\\\cite{keshavarz2020sefr,disabato2020incremental,lin2022device, lee2020learning}\\\cite{ren2021tinyol, profentzas2022minilearn,sudharsan2021ml,wulfert2024aifes}\\\cite{lee2020learning, sudharsan2021train}}&\makecell{\cite{gimenez2023embedded}}&\cite{sudharsan2021globe2train}, \textsc{RockNet}\\
    \end{tabular}
    \vspace{-0.3cm}
\end{table}

\textsc{RockNet} is an online learning method. Therefore, we focus the discussion on related works that also performs actual training. We structure the discussion as shown in Table~\ref{tab:relatedwork}: We review works on central and distributed TinyML (Section~\ref{sec:tinymlrelatedwork}), then discuss TinyML hardware implementations in Section~\ref{sec:hardwarerelatedwork} and at the end discuss networked solutions for distributed ML~\ref{sec:communicationrelatedwork}.
We focus on the high-level concepts behind these methods and details that are relevant to our work.

\subsection{TinyML Methodologies}
\label{sec:tinymlrelatedwork}
\subsubsection{Central TinyML}
\label{sec:centrallearningrelatedwork}
There are two main strategies for central TinyML.
The first focuses on fine-tuning pretrained neural networks (NN), either by training the last layer~\cite{ren2021tinyol,disabato2020incremental}, tuning parts of hidden layers~\cite{lin2022device, cai2020tinytl}, or by quantization and pruning~\cite{profentzas2022minilearn}.
However, fine-tuning requires good prior knowledge of the data the devices will encounter.
Without this, especially in high privacy scenarios, we fine-tune out-of-distribution, which can work well in some but may not achieve good accuracy in other cases~\cite{wenzel2022assaying,teney2024id}.

The second strategy involves learning without pretraining~\cite{sudharsan2021ml, keshavarz2020sefr, matsutani2022device, de2022device, de2023mu, wulfert2024aifes} resulting in suboptimal accuracy due to resource constraints.
Some works focus on training linear classifiers only~\cite{sudharsan2021ml,sudharsan2021train,keshavarz2020sefr}, 
and others train NNs~\cite{matsutani2022device, de2022device, de2023mu, wulfert2024aifes} \cite{lee2020learning} that must be small and thus have low accuracy.

\textsc{RockNet} enables training a classifier without pretraining, setting it apart from the first central TinyML strategy.
Moreover, unlike the second strategy, by utilizing special resource-efficient classifiers in conjunction with split learning, we achieve high accuracy.
Illustrating this, Section~\ref{sec:experiments} compares \textsc{RockNet} to AIfES~\cite{wulfert2024aifes}, a central TinyML method for NN training.

\subsubsection{Distributed TinyML}
\label{sec:distributedlearningrelatedwork}

As explained in Section~\ref{sec:introduction:contributions}, distributed learning has two branches: Federated and split learning.

\fakepar{Federated TinyML}
Federated learning involves each device running the entire model on its local data.
The devices repeatedly merge their models, e.g., by averaging parameters~\cite{hard2018federated, gao2020end, kopparapu2021tinyfedtl, kopparapu2022tinyfedtl,llisterri2022device,wulfert2023tinyfl,gimenez2023embedded}.
As a result, federated learning preserves privacy between devices by not sharing local data.
A disadvantage of federated learning is that each device must run the ML pipeline for the entire model, which is challenging for resource-constrained devices~\cite{thapa2022splitfed}. 

In our setup, exchanging local data between devices does not raise significant privacy concerns (cf. Section~\ref{sec:problemdefinition}); therefore, the primary advantage of federated learning is less pronounced. 
Consequently, its disadvantage, the requirement for each device to train the entire model, outweigh its benefits. 
Federated learning thus faces similar limitations as central TinyML, resulting in reduced accuracy when implemented on ultra-low-power hardware.

\fakepar{Split TinyML}
With split learning, each device calculates different parts of the model, which effectively reduces memory utilization and compute load~\cite{jin2019split,vepakomma2018split, thapa2022splitfed, gao2020end,gao2021evaluation, sudharsan2021globe2train}.
However, split learning needs significant communication~\cite{gao2021evaluation}, which can negate the benefits of parallelism by increasing training times and energy usage.
In \textsc{RockNet}, we leverage split learning to reduce resource consumption per device. 
By considering communication costs during the design of ML and utilizing an efficient communication protocol, we can effectively reduce the latency and energy of communication.

\subsection{Learning on Ultra-low-power Hardware}
\label{sec:hardwarerelatedwork}
Investigating online learning on real hardware is essential, as this allows us to operate with realistic hardware constraints, shows what is feasible with state-of-the-art algorithms and gives us true real-world performance.
Here, we focus on ultra-low-power hardware, specifically microcontrollers.
This sets us apart from studies that explore learning on more powerful embedded hardware like Raspberry Pis or smartphones~\cite{gao2020end, gao2021evaluation, hard2018federated, matsutani2022device, sudharsan2021imbal,sudharsan2021globe2train, mathur2021device, jiang2022model} or microcontrollers with specialized tensor cores~\cite{ravaglia2021tinyml} (Table~\ref{tab:relatedwork}).
These, although resource-constrained, still have significantly more resources than the ultra-low-power hardware we consider.
Current TinyML methods for such ultra-low-power hardware focus on central~\cite{sudharsan2021ml, keshavarz2020sefr, ren2021tinyol, disabato2020incremental, lin2022device, cai2020tinytl,wulfert2024aifes} and federated TinyML~\cite{gimenez2023embedded}.
Although these methods perform well within their setups, they suffer from the inherent issues of central and federated learning, as outlined above.
Split learning can solve these issues. The only existing split learning framework for ultra-low-power devices is Globe2Train~\cite{sudharsan2021globe2train}, which trains multiple base linear classifiers across multiple devices. 
However, Globe2Train relies on a central server to aggregate and distribute the training, whereas \textsc{RockNet} eliminates the need for any server-based orchestration by operating in a fully decentralized manner. 
Additionally, Globe2Train uses internet-based communication, while \textsc{RockNet} operates with purely ultra-low-power wireless communication.

To date, \textsc{RockNet} is the first work to bring decentralized split learning to ultra-low-power devices including ultra-low-power wireless communication.
We show that by leveraging split learning on constrained hardware, \textsc{RockNet} achieves state-of-the-art accuracy in a variety of setups, thus expanding the possibilities of high-accuracy machine learning for this class of devices. To highlight the performance benefits, we compare \textsc{RockNet} against AIfES\cite{wulfert2024aifes}, a state-of-the-art solution for training neural networks on a single ultra-low-power device, and demonstrate that \textsc{RockNet} significantly outperforms it in accuracy.

\subsection{Communication Solutions for Distributed Learning}
\label{sec:communicationrelatedwork}
A key component of distributed learning, whether federated or split, is the wireless communication between devices. 
Most existing approaches rely on high-power methods such as WiFi via a central router~\cite{hard2018federated, gao2020end} or WiFi mesh protocols~\cite{wulfert2023adaptive}. These solutions not only consume far more power than what ultra-low-power hardware can support (which typically uses BLE or IEEE 802.15.4) but also require bandwidth levels that such hardware cannot provide.

The only current ultra-low-power communication solution for distributed learning was proposed by Gimenez et al.~\cite{gimenez2023embedded} for ultra-low-power federated learning. Their approach employs a LoRa mesh network with a peer-to-peer protocol, though it was evaluated only on three devices with one-hop communication.

In contrast, our solution supports low-power multi-hop communication and our evaluation demonstrates a true multi-hop network with up to 20 devices. 
Rather than relying on peer-to-peer exchanges, our method implements efficient round-based all-to-all communication, closely resembling the allgather and allreduce schemes common in many ML parallelization strategies~\cite{zhou2023accelerating, du2024co}.

\section{\textsc{RockNet}: Distributed Learning on Ultra-low-power Devices}
\label{sec:method}
This section introduces \textsc{RockNet}.
We designed it to be capable of learning timeseries classification over a wireless network of ultra-low-power devices.
First, we provide an overview of our solution, followed by a detailed explanation of its components.
Afterwards, we explain how we optimize \textsc{RockNet}'s resource consumption and provide a formal analysis of its resource consumption.

\subsection{Overview}

\textsc{RockNet} enables learning in CPS solving the problem defined in Section~\ref{sec:problemdefinition}, especially under \challengelowpower{} and \challengebandwidth{}.
Based on the analysis of related work and the problem setting, we employ split learning to pool the constrained device resources.
This approach overcomes the limitations of individual devices since each device only needs to store and compute parts of the model.
Our split learning design follows the following strategy.

First, we use a suitable ML method that meets the following properties:
\begin{itemize}
  \item[\propertyefficiency] \label{pro:efficiency} \textbf{Resource Efficiency and High Accuracy.} Although we increase available compute resources through split learning, the combined resources are still severely constraint (\challengelowpower). The ML method must account for this and needs to be resource efficient, while yielding highly accurate classification results. 
  \item[\propertyparallel] \label{pro:parallel} \textbf{Efficiently Parallelizable.} As we split the ML method among devices, it needs to be parallelizable. The parallelization must be communication efficient to minimize overhead.
\end{itemize}
Subsequently, we establish a parallelization strategy for the machine learning method. 
Ultimately, we implement this strategy on distributed hardware communicating over a wireless multi-hop network. To achieve this, we present a specialized, reliable, and efficient communication protocol. 

Following this strategy, we use \textit{Random Convolutional Kernel Transform classifiers} (ROCKET)~\cite{dempster2020rocket} as ML method, because it fulfills \propertyefficiency{} and \propertyparallel{} as we explain in the following.
After designing a parallelization method for ROCKET, we derive requirements on the communication protocol.
To fulfill these requirements, we exploit recent advances in ultra-low-power wireless networking by building on a communication protocol called Mixer~\cite{Mixer} (details in Section~\ref{sec:mixer}).
Mixer's efficient and reliable all-to-all data exchange is ideal for distributing ROCKET, where input data and intermediate results must be shared among devices.
Additionally, Mixer provides network-wide time synchronization, which we exploit to efficiently parallelize ROCKET's compute.

\subsection{Foundation: ROCKET Classifiers}
\label{sec:foundation_rocket}
ROCKET~\cite{dempster2020rocket} combines efficiency with state-of-the-art accuracy (\propertyefficiency). 
In recent years, multiple derivatives of ROCKET emerged, like MiniROCKET~\cite{dempster2021minirocket}, MultiROCKET~\cite{tan2022multirocket} and HYDRA~\cite{dempster2023hydra}, which we summarize all under the term ROCKET in the following.
ROCKET algorithms have proven to be very successful in various non-linear timeseries classification tasks. 
First, the algorithms have shown competitive accuracy to other state-of-the-art classifiers like NNs, HIVE-COTE~\cite{lines2018time}, TS-CHIEF~\cite{shifaz2020ts} and ResNet on the UCR archive~\cite{UCRArchive, UCRArchive2018}, which contains 128 toy and real-world problems like ECG classification.
Thus, we refer to ROCKET as achieving state-of-the-art accuracy for timeseries classification.
Since \textsc{RockNet} represents an exact distributed implementation of ROCKET, it inherits ROCKET's state-of-the-art accuracy. This enables us to combine ultra-low-power learning on distributed hardware with cutting-edge performance for the first time.

In recent years, ROCKET has been successfully used in different applications like breast cancer classification~\cite{prinzi2024breast}, depression detection~\cite{cai2023depression}, flight performance analysis of pilots~\cite{moore2023efficient}, human activity recognition~\cite{bondugula2023novel}, astronomical seeing prediction~\cite{ni2024efficient}, activity classification for power tools~\cite{giordano2023optimizing}, and fault prediction for automated teller machines~\cite{vargas2023hybrid}. 

ROCKET not only achieves higher accuracy but is also significantly more computing efficient than state-of-the-art classifiers such as HIVE-COTE~\cite{lines2018time}, TS-CHIEF~\cite{shifaz2020ts}, and ResNet. The high computing demands of these methods have hindered the realization of high-accuracy time series classification on ultra-low-power hardware. 
We demonstrate that ROCKET, when realized via \textsc{RockNet}, overcomes these limitations.

ROCKET first transforms a timeseries into a very large feature space ($\sim 1\cdot10^5$ features).
Afterward, a linear classifier in this feature space classifies the timeseries.
The feature space transformation is based on convolutions with randomly sampled kernels.
After each convolution, ROCKET extracts features from the resulting timeseries, using the proportion of positive values (PPV)~\cite{dempster2020rocket, dempster2021minirocket, tan2022multirocket} or a dictionary method~\cite{dempster2023hydra}.
We can thus write its total behavior as 
\begin{align}
  \hat{c} =\ & h_W(x) = \text{softmax}(W f) \\ 
    =\ & \text{softmax}(W\begin{bmatrix} 1, g(k_1\ast x), g(k_2\ast x), \dots, g(k_{M-1}\ast x)\end{bmatrix}^T),\nonumber
\end{align}
where $\hat{c}\in[0,1]^{n_\mathrm{c} \times 1}$ is the predicted class probability with $n_\mathrm{c}$ being the number of classes, $f\in\realnmbr^{M\times 1}$ is the feature vector with length $F$, $M$ is the number of features, $k_i\in\realnmbr^{K_i}$ are the kernels of the convolution $k_i \ast x$ with length $K_i$ and padding $P_i$, $x\in\realnmbr^{T}$ is the input timeseries with length $T$ and $g$ is the feature extraction function, generating a scalar out of a timeseries (e.g., PPV). 
The kernels $k_i$ are sampled before training and remain constant.
Therefore, ROCKET only needs to learn the linear classifier's weights $W$, e.g., using cross-entropy loss and gradient descent~\cite{dempster2020rocket}, making the training very efficient.

ROCKET thus offers the high accuracy and computing efficiency \textsc{RockNet} needs (\propertyefficiency).
However, bringing training of ROCKET to ultra-low-power hardware is still an open challenge due to its large feature space.
The features, weights and optimizer consume up to thousands of kilobytes of RAM, too much for common ultra-low-power devices.

To date, only inference of ROCKET is achieved on ultra-low-power hardware~\cite{giordano2023optimizing}.
However, to achieve this, the approach reduces the number of features from 10,000 to 336 to fit in RAM and performs learning offline.

In contrast, \textsc{RockNet} can run ROCKET completely on ultra-low-power devices, solving \challengelowpower{}, considering not only inference but also \emph{training} of ROCKET on ultra-low-power hardware for the first time.
It achieves this through split learning, distributing features and weights onto multiple devices.
For this, \textsc{RockNet} exploits that we can efficiently parallelize ROCKET (\propertyparallel{}) as we now demonstrate.

\subsection{\textsc{RockNet}: Learning Architecture}

\begin{figure*}[t]
  \centering
  \includegraphics[width=0.9\linewidth]{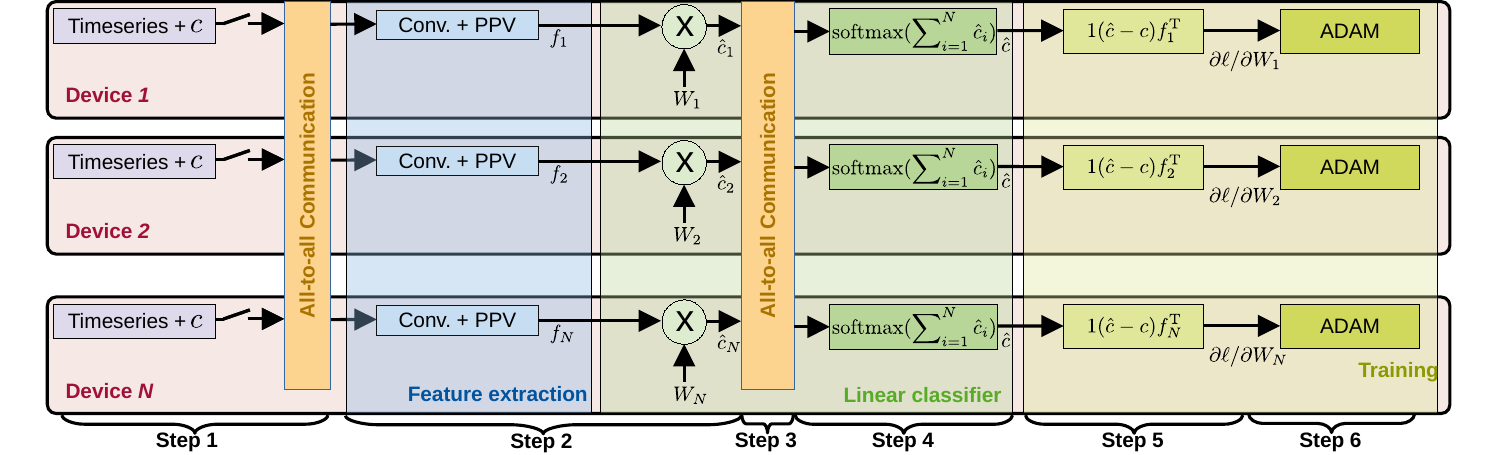}
  \vspace{-0.4cm}
  \caption{Overview of \textsc{RockNet}. \textit{During the forward pass (feature extraction + linear classifier), all devices exchange their data with each other. The training step does not require any additional communication, making \textsc{RockNet} more efficient.}}
  \label{fig:distributedrocketoverview}
  \vspace{-0.4cm}
\end{figure*}

The goal of \textsc{RockNet} is to perform the exact computations as ROCKET to achieve identical accuracy.
We split ROCKET along the feature dimension.
Each device $i$ calculates different features $f_{i}$ in $f$ (Fig.~\ref{fig:distributedrocketoverview})
\begin{equation}
  f = \begin{bmatrix} f_{1}^T, f_{2}^T, \dots, f_{N}^T \end{bmatrix}^T.
\end{equation}

Two strategies exist to partition the linear classifier: input and output partitioning~\cite{stahl2021deeperthings}. With output partitioning, devices first share their feature vectors, multiply them with the corresponding values from $W$, and then exchange the resulting products. This process requires transferring the entire feature vectors (e.g., 10,000 float values) and performing a second communication step, leading to excessive overhead.

In contrast, input partitioning allows each device to multiply its features by the corresponding values from $W$ locally before exchanging the partial results and summing them. As a result, we need to exchange only $N n_\mathrm{c}$ values. For example, with 20 devices and 10 classes, this approach reduces communication by a factor of 50 compared to output partitioning. The only trade-off is a redundant calculation of the sum of the partial results on every device, which requires an additional $(N-1)n_\mathrm{c}$ float additions per device. However, these few hundred extra operations are far less expensive than the communication overhead of output partitioning.
We hence partition the linear classifier using input partitioning.
Each device $i$ is assigned a portion $W_{i}$ of the weights~$W$
\begin{equation}
  W = \begin{bmatrix} W_{1}, W_{2}, \dots, W_{N} \end{bmatrix}.
\end{equation}
This reduces overall computing time and decreases RAM usage since each device has to store fewer features and weights.

All devices process one timeseries simultaneously.
First, they predict the class $\hat{c}$ of the timeseries $x$ (forward pass).
Then, they calculate the gradient of the cross-entropy loss between their prediction and the true class label $c$ with respect to the weights $W$ (backward pass). 
The devices accumulate the gradient over multiple timeseries and update the weights once the batch size is reached, using ADAM~\cite{kingma2014adam}.
Evaluation steps can be performed between training steps to stop early and thus avoid overfitting.
In detail, all devices
perform the following steps simultaneously (see Figure~\ref{fig:distributedrocketoverview}).

\fakepar{Step 1: Measuring Data} One device $j$ sends its timeseries $x$ with class $c$ from its local data $\mathcal{D}_j$ to all devices.
  \footnote{The original ROCKET methods only support univariate timeseries.
  While we stick with the original univariate setting for reasons of comparability, \textsc{RockNet} in principle also supports multivariate extensions of ROCKET.}

  \fakepar{Step 2: Feature Extraction and Local Computation} Each device~$i$ extracts its assigned features $f_{i}$ from the timeseries and multiplies these features $f_{i}$ with its portion of $W_{i}$.

  \fakepar{Step 3: All-to-all Exchange of Results} The results $\hat{c}_{i} = W_{i}f_{i}$ are exchanged via all-to-all communication, where all devices send their $\hat{c}_{i}$ values and receive the $\hat{c}_{i}$ values from all other devices.

  \fakepar{Step 4: Summation of Results}
  Each device sums the received values and applies the softmax operation to obtain class probabilities for the input $x$. As a result, each device knows the predicted class $c$. \textsc{RockNet} performs the same computations as ROCKET:
  \begin{align}
    \text{softmax}(\sum_{i=1}^{N}W_{i}f_{i}) & =  \text{softmax}\big(\begin{bmatrix} W_{1}, W_{2}, \dots, W_{M} \end{bmatrix}\begin{bmatrix} f_{1}^T, f_{2}^T, \dots, f_{N}^T \end{bmatrix}^T\big)\nonumber\\ 
    & = \text{softmax}(W\begin{bmatrix} 1, g(k_1\ast x), g(k_2\ast x), \dots, g(k_{M-1}\ast x) \end{bmatrix}^T).\nonumber
  \end{align} 

  \fakepar{Step 5: Backward Pass} Up to this step, the devices have completed the forward pass, i.e., inference.
  At this point, each device~$i$ already has all the information needed to perform the backward step without additional communication, making the parallelized training process very communication efficient (\propertyparallel).
  Each device calculates the gradient of the cross-entropy loss $l$ between its prediction $\hat{c}$ and the true value $c$ with respect to its weights $W_{i}$
  \begin{equation}
    \frac{\partial}{\partial W_{i}}l(\hat{c}, c) = (\hat{c} - c)f_{ i}^T.
  \end{equation}

  \fakepar{Step 6: Update Step} After accumulating gradients over multiple timeseries, each device~$i$ updates its weights $W_{i}$ using ADAM~\cite{kingma2014adam}. To achieve this, the first and second moment estimates of the associated weights must be retained in memory. To minimize memory usage, we apply quantization as outlined in Section~\ref{sec:optimizing_execution:quantization}. Since each device possesses all necessary information for an ADAM update of its respective weights, it can execute this operation locally. Consequently, Steps 5 and 6 collectively conduct an exact gradient descent via ADAM, ensuring that all theoretical insights and empirical strategies of ADAM remain applicable.

  To summarize, \textsc{RockNet} can efficiently distribute the training of ROCKET without much communication (\propertyparallel).
  It communicates the timeseries, usually a few hundred of bytes, and the parts of the linear classifier $\hat{c}_i$, a few tens of bytes, for the forward pass.
  For the backward pass, \textsc{RockNet} needs no additional communication.

\subsection{\textsc{RockNet}: Wireless Communication Support}
\label{sec:mixer}

The \textsc{RockNet} parallelization strategy demands fully connected and synchronized communication, a requirement that sharply contrasts with the sparse and dynamic characteristics of the multi-hop network used for device communication. To reconcile these differences, we utilize a tailored multi-hop communication protocol that must meet the following requirements:
\begin{itemize}
\item[\requirementalltoall] \label{req:alltoall} \textbf{All-to-All Data Exchange.} Each device must transmit its data and the results to all other devices (Steps 1 and 3), necessitating all-to-all data exchange.
\item[\requirementreliable] \label{req:reliable} \textbf{Reliable and Efficient Communication.} Data exchange must be highly reliable with negligible message losses, as any lost message hinders learning progress.
Although we efficiently parallelize ROCKET,
the data exchange must also be efficient to close the communication bottleneck further. 
\item[\requirementsync] \label{req:sync} \textbf{Synchronized Execution.} \textsc{RockNet}'s parallelization strategy requires a globally synchronized execution of computation and communication to minimize waiting times and accelerate the learning process.
\end{itemize}

Meeting requirements \requirementalltoall{}, \requirementreliable{}, and \requirementsync{} are challenging as
wireless communication is notoriously unreliable because signals are exposed to environmental disturbances like interference and fading effects.   
The severely limited communication bandwidth and energy availability (\challengebandwidth{}) further add to these challenges.
These issues are further amplified by the sparsely connected topology of the multi-hop network, while device mobility introduces a dynamic network structure that reinforces these challenges.

To address these key challenges and satisfy \requirementalltoall{}--\requirementsync{}, \textsc{RockNet} leverages a state-of-the-art low-power wireless communication protocol called Mixer~\cite{Mixer}.
Mixer combines synchronous transmissions~\cite{Glossy} and random linear network coding~\cite{Ho2006}, enabling more efficient and reliable data exchange compared to traditional wireless protocols~\cite{Schuss2017,zimmerling2020}.
The effectiveness of similar synchronous transmissions-based protocols has already been demonstrated for various real-world applications, including wireless feedback control with dependability and real-time guarantees~\cite{mager2019feedback, Trobinger2021}.
 
Mixer organizes its operation into discrete rounds of deterministic and constant duration, each of which is subdivided into multiple slots synchronized with a precision of a few microseconds. 
During any given slot, a device operates in either transmit or receive mode, with the decision governed by heuristics (for further details, see~\cite{Mixer}). 
When transmitting, a device sends a linear combination of its own message and the data received from other devices in previous slots, a process known as network coding~\cite{Ho2006}. 
Each device stores the data it receives in memory, up to the number of transmitted messages. 
This stored data represents linear combinations of the messages originally sent by all devices. 
By solving the resulting system of linear equations at the end of the round, the device can decode every original message, thus realizing an all-to-all communication pattern (\requirementalltoall{}).

When multiple neighboring devices transmit simultaneously, a receiving device exploits the capture effect~\cite{Leentvaar1976}, which is the underlying principle of synchronous transmission techniques~\cite{Glossy}.
Owing to this effect, the receiver can extract the signal with the highest strength with high probability, thereby eliminating the need for complex scheduling mechanisms to avoid simultaneous transmissions.
This reliance on the capture effect hence significantly reduces communication overhead making communication very efficient.

Achieving this benefit, however, requires that slots be synchronized with an accuracy of a few microseconds, a challenging task due to inevitable clock drifts. 
Mixer effectively overcomes this challenge by enabling each receiving device to adjust its clock “on the fly” to that of the transmitting device using a phase-locked loop that uses the precise reception time of the signal. 
This dynamic synchronization satisfies \requirementsync{} efficiently.

This combination of synchronous transmissions and network coding leads to a very high performance and reliability of Mixer~(\requirementreliable{}).
In experimental evaluations, Mixer exhibited a message loss rate below \SI{0.01}{\percent} even under conditions of rapid device movement~\cite{Mixer}.
Furthermore, Mixer achieves order-optimal scaling of $\mathcal{O}(N+T)$, where $T$ scales linearly with the message size, thereby ensuring minimal communication latencies.

This brief overview only scratches the surface of Mixer's internal mechanisms—numerous other aspects contribute to its beneficial properties. 
For an in-depth explanation, we refer to the original Mixer paper\cite{Mixer}.

Additional to all of this, Mixer has another benefit.
It spares the split learning side of \textsc{RockNet} from having to manage the intricate dynamics of wireless multi-hop communication.
Mixer functions like a conductor, synchronizing all devices into rounds of all-to-all communication (\requirementalltoall{} and \requirementsync{}) adding an abstraction layer over the multi-hop network.

Within this abstraction layer, communication via Mixer emulates a wired bus that also provides a global clock signal for all devices.
Synchronized by this clock, all devices simultaneously write their messages to the bus, and after a fixed, deterministic latency, every device receives all the messages. 
Following this, each device is allocated a constant duration to perform its computations. Once this computing phase concludes, the devices write their data to the bus again, and the cycle continues. 
This schedule aligns exactly with the timing requirements of our parallelization strategy (cf. Fig.~\ref{fig:distributedrocketoverview}).

Overall, Mixer's combination of synchronous transmissions and network coding offers communication properties that are well-suited for the dynamic nature of various CPS in general and for the specific requirements of split learning in \textsc{RockNet} in particular.

\begin{figure*}[t]
  \centering
  \includegraphics[width=0.9\linewidth]{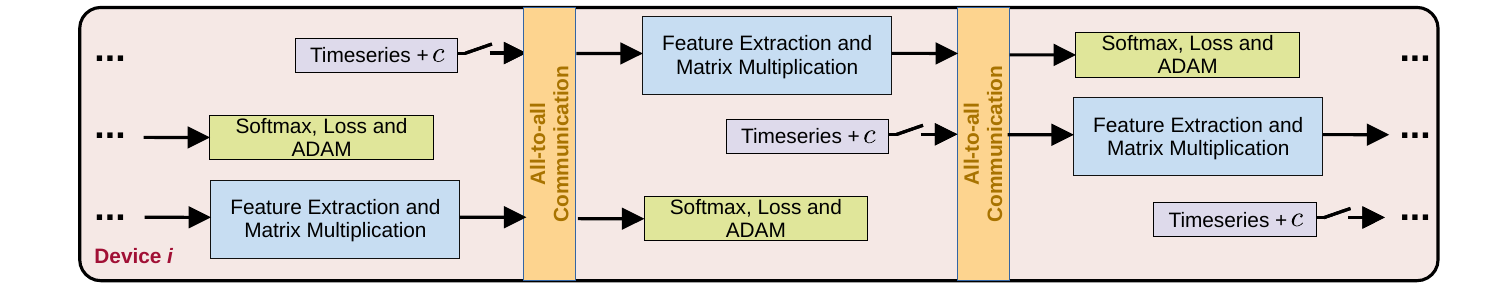}
  \vspace{-0.4cm}
  \caption{Pipelining of \textsc{RockNet}. \textit{In our implementation, we process three timeseries simultaneously to reduce the overhead induced by communication.}}
  \label{fig:distributedrocketpipeline}
  \vspace{-0.4cm}
\end{figure*}

\subsection{\textsc{RockNet}: Optimizing the Execution}
\label{sec:optimizing_execution}
Despite \textsc{RockNet}'s inherent resource efficiency, we further enhance it through pipelining and quantization.

\subsubsection{Pipelining} 
Mixer's communication latency has a constant offset due to overhead like synchronization and computings, independent of data size. 
During one \textsc{RockNet} round, this overhead would occur twice (in Steps 1 and 3), which is particularly significant in Step 3, where the transmitted messages are small (only one \qty{32}{\bit} floating point value per class).
To speed up \textsc{RockNet}, we use pipelining by combining the execution of Step 1 for the next data, the execution of Steps 2 and 3 for the current data and the execution of Steps 4, 5 and 6 for the previous data into a single round (Figure~\ref{fig:distributedrocketpipeline}).
This merges the communication rounds of Steps 1 and 3 into one. 
Thus, we need only $n+1$ instead of $2n$ communication rounds for $n$ timeseries, reducing latency caused by communication overhead.

\subsubsection{Quantization} 
\label{sec:optimizing_execution:quantization}
With quantization of data, we aim to reduce both communication latency and memory.

\fakepar{Quantization of Timeseries}
To reduce communication latency, we minimize the amount of transmitted data by quantizing \qty{32}{\bit} floating point values to \qty{8}{\bit} integers. 
We specifically focus on quantizing the timeseries, which constitute a significant portion of the transmitted data, measuring in hundreds of bytes, compared to only four bytes per class consumed by $\hat{c}_i$.
The quantization uniformly maps the range of values onto \qty{8}{\bit}~\cite{gholami2022survey,wu2020integer}, efficiently reducing the amount of data transmitted by up to \SI{75}{\percent}.

\fakepar{Quantization of ADAM} 
ADAM, used for training, consumes three times the memory of the weights: once for the gradients and twice for its internal states (first and second-order momentum), significantly contributing to memory consumption.
Reducing the overall memory consumption of \textsc{RockNet} by about \SI{30}{\percent}, we follow Dettmers et al.~\cite{dettmers20228bitadam} and quantize ADAM's internal states to \qty{8}{\bit}.

8-bit ADAM uses a non-uniform quantization that provides a higher accuracy for small values than uniform quantization.
Additionally, it uses block-wise quantization.
The state is split into blocks (here, of size 256), for which its own scaling is calculated.
This reduces quantization errors and also memory during the execution of ADAM, as we can buffer it into chunks of 256 values.

\subsection{Analysis of Resource Consumption}
\label{sec:analysis_res_consumption}
This section provides a formal analysis of the resource consumption associated with our approach, focusing on three key metrics: the amount of memory used, the number of floating point operations performed, and the volume of data transmitted per round. 
In the subsequent Section~\ref{sec:experiments}, we experimentally assess resource consumption, particularly in terms of latency and energy usage, through hardware experiments.

\subsubsection{Memory}
\label{sec:analysis_res_consumption:memory}
The memory consumption (RAM) of a device during each round is quantified in bytes, under the assumption that the number of features $F$ is divisible by the number of devices $N$.
It can be expressed as follows
\begin{align}
  \text{memory} =\ & \underbrace{4n_\mathrm{c}\frac{F}{N}}_{\text{Weight}} 
                   + \underbrace{4n_\mathrm{c}\frac{F}{N}}_{\text{Gradient}} 
                   + \underbrace{2n_\mathrm{c}\frac{F}{N}}_{\text{ADAM states}} 
                   + \underbrace{4\frac{F}{N}}_{\text{Features}}
                   + \underbrace{T}_{\text{Timeseries}}
                   + \underbrace{C_\mathrm{memory}}_{\text{Rest}},
\end{align}
where $C_\mathrm{memory}$ represents overhead components such as smaller buffers and intermediate variables, which in our implementation amount to approximately \qty{25}{\kilo\byte}.
Consequently, the complexity of memory consumption regarding number of devices is characterized by $O(\frac{1}{N}+T+C_\mathrm{memory})$.
As a result, increasing the network size results in a decrease in memory that converges to a constant overhead.

\subsubsection{Floating Point Operations}
\label{sec:analysis_res_consumption:flops}
For floating point operations, we do not differentiate between operations such as addition, multiplication, multiply-and-accumulate, and division. Operations performed during the Mixer phase are excluded from consideration as they are difficult to assess.
The number of operations per round is given by the following expression (assuming for simplicity that all convolutions have the same length $K$ and padding $P$)
\begin{align}
  \text{ops} & = \underbrace{K\frac{F-K+2P+1}{N}}_{\text{Convolution}} 
                   + \underbrace{\frac{F-K+2P+1}{N}}_{\text{Feature Extraction}} 
                   + \underbrace{n_\mathrm{c}\frac{F}{N}}_{\text{Weight Multiplication}} 
                   + \underbrace{c_\mathrm{exp}n_\mathrm{c}+2n_\mathrm{c}-1}_{\text{Softmax}}\nonumber\\
                   &\opindent+ \underbrace{n_\mathrm{c} + n_\mathrm{c}\frac{F}{N}}_{\text{Gradient}}
                   + \underbrace{(14+c_\mathrm{sqrt})n_\mathrm{c}\frac{F}{N}}_{ADAM}
                   + \underbrace{C_\mathrm{ops}}_{\text{Rest}}
\end{align}
where $c_\mathrm{exp}$ and $c_\mathrm{sqrt}$ are floating operations per calculation required for executing a single exponential or square-root function, respectively, and $C_\mathrm{ops}$ accounts minor overheads.
Thus, the complexity of floating point operations concerning the number of devices is analogous to memory consumption and can be expressed as $O(\frac{1}{N}+C)$.
Increasing the network size hence reduces the computational load.

\subsubsection{Communication} 
We evaluate the data communication as the sum of data each device intends to transmit.
Due to the Mixer mechanism, where each device combines messages and transmits multiple times across various slots (see Section~\ref{sec:mixer}), the volume of data sent via the wireless channel is higher. 
The precise relationship between these factors is challenging to determine, as it depends on the configuration of the multi-hop network (cf. the analysis of Hermann et al.~\cite{Mixer}). 
To address this, we perform an empirical analysis in Section~\ref{sec:scaling}.

The amount of data communicated is in byte
\begin{align}
  \text{data} =\ & \underbrace{T}_{\text{Timeseries}} 
                   + \underbrace{4n_\mathrm{c}N}_{\text{Prediction}}.
\end{align}
Thus, its complexity is $O(N+T)$.
Consequently, the amount of data increases steadily with each additional device, in addition to a constant offset.

This section has provided an analytical examination of memory complexity, floating-point operations, and communication. 
In the following section, we will illustrate how these properties apply to real hardware implementations.

\section{Experimental Results}
\label{sec:experiments}
This section studies the performance and efficiency of \textsc{RockNet} using simulations and measurements from a wireless testbed with 20 ultra-low-power devices.
Our key findings are as follows:
\begin{itemize}
  \item \textsc{RockNet} can train ROCKET classifiers on severe memory constraint ultra-low-power hardware. While running ROCKET on just one device would exceed its RAM by \SI{428}{\percent}, \textsc{RockNet} running on 20 devices utilizes \SI{37.2}{\percent} of the RAM and reduces latency and per-device energy by up to \SI{89}{\percent} compared to ROCKET running on one device. This is the first demonstration of training of SOTA timeseries classifiers on ultra-low-power hardware.
	\item \textsc{RockNet}'s accuracy surpasses central on-device NN training~\cite{wulfert2024aifes} over time in \SI{78.51}{\percent} of runs on the UCR archive with a mean absolute accuracy improvement of \SI{12.7}{\percent}.
	\item There is a price to pay for the improved accuracy: Longer training times (up to $86\times$) and higher energy consumption (up to $57\times$) compared to central on-device NN training~\cite{wulfert2024aifes}.
	\item \textsc{RockNet} effectively pools the resources of multiple devices.
  Training times and RAM utilization decrease by up to \SI{65.0}{\percent} and \SI{55}{\percent} when \textsc{RockNet} scales from five to 20 devices. 
\end{itemize}

\subsection{Methodology}
\begin{figure}[t]
  \centering
  \includegraphics[width=0.4\linewidth]{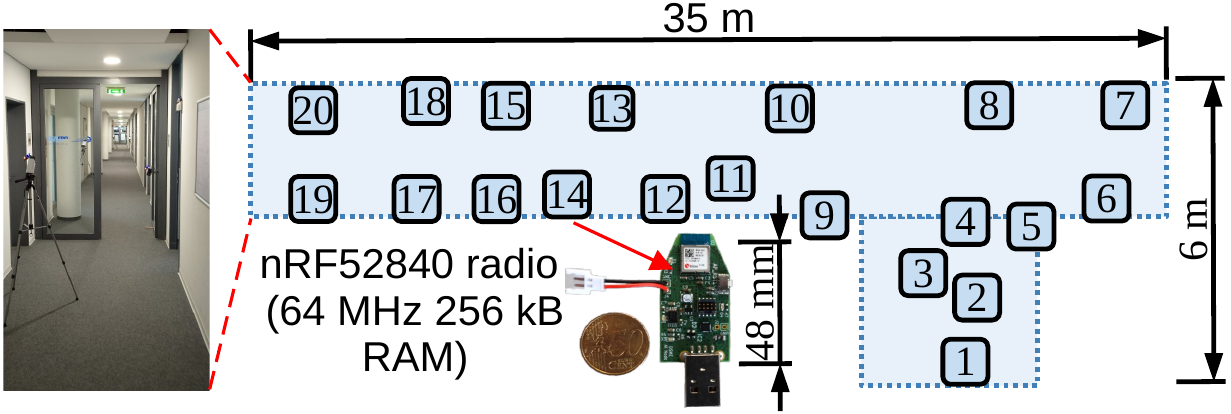}
  \vspace{-0.3cm}
  \caption{Overview of our testbed. \capt{It consists of up to 20 ultra-low-power Bluetooth radios forming a two-hop network.}}
  \label{fig:testbed}
  \vspace{-0.5cm}
\end{figure}
 
\fakepar{Testbed}
Figure~\ref{fig:testbed} shows our testbed and the positions of the 20 nRF52840 devices distributed across a hallway and an adjacent office.
Each device has \SI{256}{\kilo\byte} of RAM and a 32-bit ARM Cortex-M4 ultra-low-power microcontroller running at \SI{64}{\mega\hertz}.
Devices transmit at \SI{8}{\dBm} using Bluetooth Low Energy radios in a genuine multi-hop network with at least two hops.
Interference from co-located Wi-Fi networks, a microwave, and people cater to realistic wireless conditions.

\fakepar{Datasets}
We train non-linear classifiers using labeled datasets from the popular UCR archive~\cite{UCRArchive2018,UCRArchive} containing 128 timeseries classification problems.
In our simulations, we use all 128 datasets.
For our testbed experiments, we select 3 representative CPS datasets from the UCR archive to keep the running time of the experiments manageable: \emph{OSULeaf} (nature observation, 6 classes), \emph{ElectricDevices} (smart grids, 7 classes), and \emph{FaceAll} (face detection, 13 classes). 

Datasets are shuffled and partitioned among devices, leading to identically and independently distributed (i.i.d.) local data and, consequently, i.i.d. batches during training. While this is a simplification, it closely mirrors real-world data distributions in \textsc{RockNet}. 
Devices collect data over an extended time horizon and store it locally in $\mathcal{D}_i$. 
By randomly selecting entries from $\mathcal{D}_i$ for transmission and subsequent training, the data stream from each device becomes i.i.d. 
Moreover, as each device transmits periodically (in a round-robin fashion), the batches incorporate data from all devices and are thus also approximately i.i.d.
We investigated this in Appendix~\ref{app:iid} demonstrating that non-i.i.d. data does not reduce \textsc{RockNet}'s accuracy.

\fakepar{Metrics}
We measure performance in terms of \emph{accuracy}, the percentage of correctly classified timeseries when assessing the trained models on the test datasets.
To quantify efficiency, we measure \emph{memory consumption} as the amount of RAM used, \emph{latency} as the required training time, and \emph{energy consumption} as the energy consumed during training.
We measure energy during training by logging the times a device computes, communicates, and spends in a low-power mode and multiplying these times by the corresponding average power draws.

Acknowledging the limited memory and energy budgets inherent in our setting, we report the per-device memory and energy consumption. Analyzing how these resources scale, particularly with respect to the number of participating devices, provides valuable insights into the conditions under which the investigated algorithms are inside the constraints and become feasible.

\fakepar{Approaches}
Our \textsc{RockNet} implementation uses MiniROCKET, the most memory and computing efficient ROCKET variant that achieves the same accuracy as the original ROCKET~\cite{dempster2021minirocket}.

We compare \textsc{RockNet} to two state-of-the-art baselines:
\begin{itemize}
 \item \emph{AIfES} is a very recent TinyML approach for NN training on a single ultra-low-power device without pretraining~\cite{wulfert2024aifes}. To ensure a fair comparison, we tune AIfES's NN to use as much as possible of the \SI{256}{\kilo\byte} of RAM available on the nRF52840 without causing divergence issues for any of the 128 datasets. We thus conduct all experiments using ten hidden layers with 17 neurons each, which consumes \SI{74}{\percent} of the available RAM. As for \textsc{RockNet}, the ADAM optimizer is used to train the NN.
 \textit{Remark:} The network architecture is constrained by what is currently supported by AIfES. In its original publications~\cite{dempster2020rocket,dempster2021minirocket,dempster2021minirocket}, ROCKET has been shown to outperform more complex architectures such as ResNet while requiring far fewer computing resources. Therefore, if full neural network training on MCUs were feasible, ROCKET would be expected to show a performance advantage.
\textit{Remark:} The network architecture is constrained by the current limitations of microcontroller neural network training. In its original publications~\cite{dempster2020rocket,dempster2021minirocket,dempster2021minirocket}, ROCKET has been shown to outperform more complex architectures like ResNet while requiring far fewer computing resources. Therefore, if full neural network training on MCUs were feasible, it is expected that ROCKET would show a performance advantage.
 \item \emph{Central ROCKET} refers to running MiniROCKET on a single ultra-low-power device. Because of RAM limitations, however, central ROCKET is not feasible on the nRF52840 (cf.~Section~\ref{sec:scaling}). To still be able to compare against this state-of-the-art baseline, we estimate all metrics based on \textsc{RockNet}'s measurements. Since \textsc{RockNet} is an exact distributed realization of ROCKET, both approaches achieve the same accuracy. To determine central ROCKET's efficiency, we scale the computing resources of \textsc{RockNet} by the number of devices.  
\end{itemize}
For both baselines, we neglect the time and energy needed to communicate the local data $\mathcal{D}_i$ to the central device for training (see Figure~\ref{fig:setup}).
This makes our comparisons favorable to the baselines.

\fakepar{Hyperparameters}
We intentionally did not perform a dedicated hyperparameter search for \textsc{RockNet}, as it targets ultra-low-power hardware. 
In this setting, executing multiple training cycles for hyperparameter tuning is compute expensive and may exceed available budget constraints, thus necessitating the reliance on a default learning rate.
Consequently, all our experiments are carried out using a learning rate of $1 \times 10^{-3}$, which is consistent with the learning rate employed in the original ROCKET papers. Additionally, we fixed the batch size at 128.

The code used to generate the results of our experiments can be found in \href{https://github.com/Data-Science-in-Mechanical-Engineering/RockNet}{github.com/Data-Science-in-Mechanical-Engineering/RockNet}.

\subsection{Accuracy}

\begin{figure}[t] %{r}{0.25\textwidth}
  \fontsize{\plotfontsize}{\plotfontsize}\selectfont
  \centering
  \begin{subfigure}[b]{0.48\linewidth}
    \centering
    \begin{tikzpicture}[spy using outlines={circle, magnification=3, size=2cm, connect spies, every spy on node/.append style={thick}}]
    \begin{axis}[
      % at=(c), 
      % xshift=1mm,
      % anchor=north,
      xmax=100.01, xmin=-0.01,
      ymax=100.01, ymin=-0.01,
      name=plot3, xlabel={AIfES accuracy (\SI{}{\percent})}, 
      ylabel={{\textsc{RockNet}\\accuracy (\SI{}{\percent})}},
      ylabel style={align=center,yshift=-1mm},
      xlabel style={align=center,yshift=1mm, xshift=0mm},
      height=0.5\linewidth, width=0.5\linewidth,
      clip mode=individual,
      view={0}{90},
      ]

      \addplot3[scatter,mark=*, mark size=0.5pt, only marks] table [x=accNN, y=accRocket, z=distanceBoundary, col sep=comma] {images/ComparisonNNROCKET.csv};

      \addplot[color=black, line width=1pt, mark=none, dotted, dash pattern=on 1pt off 1pt] coordinates {(0,0) (100,100)};
    \end{axis}
  \end{tikzpicture}
  % \vspace{-0.3cm}
  \caption{32\,bit ADAM}
  \label{fig:32bit}
\end{subfigure}
\begin{subfigure}[b]{0.48\linewidth}
  \centering
  \begin{tikzpicture}[spy using outlines={circle, magnification=3, size=2cm, connect spies, every spy on node/.append style={thick}}]
    \begin{axis}[
      % at=(c), 
      % xshift=1mm,
      % anchor=north,
      xmax=100.01, xmin=-0.01,
      ymax=100.01, ymin=-0.01,
      name=plot3, xlabel={AIfES accuracy (\SI{}{\percent})}, 
      ylabel={{\textsc{RockNet}\\accuracy (\SI{}{\percent})}},
      ylabel style={align=center,yshift=-1mm},
      xlabel style={align=center,yshift=1mm, xshift=0mm},
      height=0.5\linewidth, width=0.5\linewidth,
      clip mode=individual,
      view={0}{90},
      ]

      \addplot3[scatter,mark=*, mark size=0.5pt, only marks] table [x=accNN, y=accRocket, z=distanceBoundary, col sep=comma] {images/ComparisonNNROCKET8bit.csv};

      \addplot[color=black, line width=1pt, mark=none, dotted, dash pattern=on 1pt off 1pt] coordinates {(0,0) (100,100)};
    \end{axis}
  \end{tikzpicture}
  % \vspace{-0.3cm}
  \caption{8\,bit ADAM}
  \label{fig:8bit}
\end{subfigure}
% \vspace{-0.4cm}
\caption{Accuracy of \textsc{RockNet} versus central NN training with AIfES. \textit{RockNet achieves a higher accuracy than AIfES in more than \SI{78}{\percent} of the runs with \SI{32}{\bit} ADAM and in \SI{67}{\percent} of the runs with \SI{8}{\bit} ADAM.}}
\label{fig:simulationresults}
% \vspace{-8mm}
\end{figure}
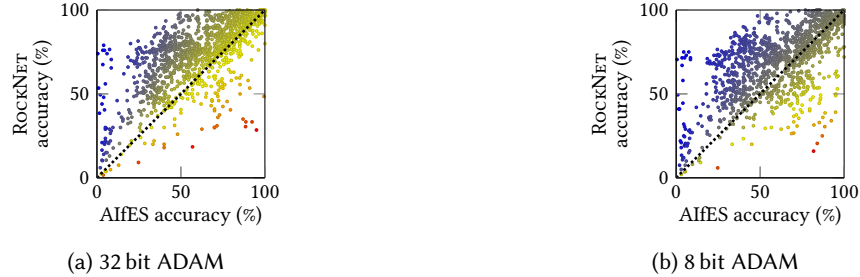

\begin{table*}[t]
  \caption{\textsc{RockNet} versus AIfES in three datasets from the UCR archive~\cite{UCRArchive2018}.}
  \label{tab:hardwareexperiments}
  \fontsize{8pt}{8pt}\selectfont
	\centering
    \ra{1.4}
    \begin{tabular}{@{}cc||rrrrr@{}}
      Dataset & Method & \makecell{Maximum\\test accuracy}& \makecell{Time to\\maximum accuracy}&\makecell{Energy per device\\to maximum accuracy}&\makecell{Time to\\surpass AIfES}&\makecell{Energy per device \\to surpass AIfES}\\
      \hline\hline
      \multirow{2}{*}{OSULeaf}&AIfES&\SI{35.5}{\percent}&\SI{0.074}{\hour}&\SI{2.970}{\joule}&-&-\\
      &\textsc{RockNet}&\SI{92.1}{\percent}&\SI{6.282}{\hour}&\SI{144.52}{\joule}&\SI{0.306}{\hour}&\SI{8.983}{\joule}\\\hline
      \multirow{2}{*}{ElectricDevices}&AIfES&\SI{51.5}{\percent}&\SI{0.100}{\hour}&\SI{4.007}{\joule}&-&-\\
      &\textsc{RockNet}&\SI{67.5}{\percent}&\SI{7.565}{\hour}&\SI{230.816}{\joule}&\SI{0.265}{\hour}&\SI{6.615}{\joule}\\\hline
      \multirow{2}{*}{FaceAll}&AIfES&\SI{47.5}{\percent}&\SI{0.132}{\hour}&\SI{5.286}{\joule}&-&-\\
      &\textsc{RockNet}&\SI{72.5}{\percent}&\SI{4.963}{\hour}&\SI{119.454}{\joule}&\SI{0.652}{\hour}&\SI{14.622}{\joule}\\
    
    \end{tabular}
\end{table*}

We begin by comparing the accuracy of \textsc{RockNet} to AIfES. Note that \textsc{RockNet} and central ROCKET have the same accuracy.

\fakepar{Settings}
We train AIfES and \textsc{RockNet} in simulation on the entire UCR archive using ten random seeds.
We randomly initialize the weights and shuffle the training and evaluation splits randomly for each seed.
We simulate \textsc{RockNet} with \SI{32}{\bit} ADAM and with \SI{8}{\bit} ADAM.
The training runs for 1000 epochs, with the model from the epoch with the highest evaluation accuracy being selected for each seed.
These models are then evaluated on the test datasets.

\fakepar{Results}
Figure~\ref{fig:simulationresults} plots, for each seed and test dataset, the accuracy of \textsc{RockNet} against the accuracy of AIfES.
We find that for \SI{32}{\bit} ADAM (Figure~\ref{fig:32bit}), \textsc{RockNet} achieves a higher accuracy than AIfES in \SI{78.5}{\percent} of the runs and a mean accuracy improvement of \SI{12.7}{\percent}.
For \SI{8}{\bit} ADAM (Figure~\ref{fig:8bit}), \textsc{RockNet} outperforms AIfES in \SI{67}{\percent} of the runs with a mean accuracy improvement of \SI{6.5}{\percent}.
While \SI{8}{\bit} ADAM saves about \SI{30}{\percent} of RAM compared to \SI{32}{\bit} ADAM, it leads to an average drop in accuracy of \SI{4.2}{\percent} in \SI{88.3}{\percent} of the runs. 

These results support our design decision to use ROCKET as the base ML technique of \textsc{RockNet}.
Thanks to \textsc{RockNet}'s split learning architecture, it outperforms state-of-the-art TinyML for NNs on ultra-low-power hardware.
The choice between \SI{32}{\bit} ADAM and \SI{8}{\bit} ADAM is a trade-off between memory consumption and accuracy.
While \SI{30}{\percent} less memory is already significant, we show in the next section that orders of magnitude higher per-device memory savings are enabled by \textsc{RockNet} \emph{without} sacrificing on accuracy.

\subsection{Efficiency}
This section evaluates the efficiency of the learning process in terms of memory consumption, latency, and energy consumption.

\fakepar{Settings}
We run experiments on our 20-node wireless testbed (Figure~\ref{fig:testbed}).
We first compare \textsc{RockNet} to AIfES on the three selected datasets: OSULeaf, ElectricDevices, and FaceAll.
Then, we analyze \textsc{RockNet}'s scaling behavior with the number of devices.
For all three datasets, \textsc{RockNet} uses \SI{8}{\bit} ADAM, which achieves the same accuracy on these specific datasets as \SI{32}{\bit} ADAM while reducing memory by about \SI{30}{\percent}. 
However, this reduction in memory consumption comes with a \SI{26}{\percent} increase in latency and energy consumption due to additional computations induced by the 8-bit dynamic tree quantization, which the hardware of the processor does not support.
This increase does not influence the general scaling behavior, so we only show results for \SI{8}{\bit} ADAM in Section~\ref{sec:scaling}.

\subsubsection{\textsc{RockNet} versus AIfES} 
\label{sec:aifesvsrocknet}

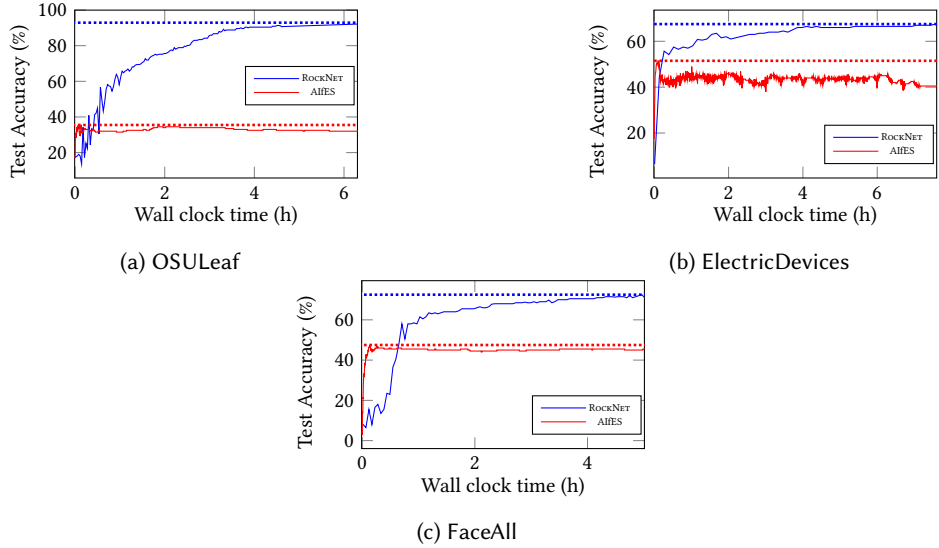
\begin{figure}[t]
  \fontsize{\plotfontsize}{\plotfontsize}\selectfont
  \centering
  \newcommand{\plotheight}{0.5\textwidth}
  \newcommand{\plotwidth}{0.7\textwidth}
  \begin{subfigure}[t]{0.48\linewidth}
  \centering
  \begin{tikzpicture}[spy using outlines={circle, magnification=3, size=2cm, connect spies, every spy on node/.append style={thick}}]

    \begin{axis}[
      xmax=6.3, xmin=-0.01,
      y label style={xshift=-1mm,yshift=\ylabelyshift},
      x label style={yshift=\xlabelyshift},
      name=plot3, 
      xlabel={Wall clock time (\si{\hour})}, 
      ylabel={{Test Accuracy (\si{\percent})}},
      height=\plotheight, width=\plotwidth,
      clip mode=individual,
      legend style={at={(0.99, 0.55)},anchor=east,nodes={scale=0.5, transform shape}}, 
      ]

      \addplot[mark=none, color=blue] table [x=timestamp, y=accuracy, col sep=comma] {images/AccuracyFinalOSULeaf20True.csv};
      \addlegendentry{\textsc{RockNet}}
      
      \addplot[color=blue, line width=1pt, mark=none, dotted, dash pattern=on 1pt off 1pt, forget plot] coordinates {(-1,92.9) (11,92.9)};

      \addplot[mark=none, color=red] table [x=timestamp, y=accuracy, col sep=comma] {images/AccuracyFinalOSULeafNN1True.csv};
      \addlegendentry{AIfES}
      \addplot[color=red, line width=1pt, mark=none, dotted, dash pattern=on 1pt off 1pt, forget plot] coordinates {(-1,35.5) (11,35.5)};
    \end{axis}
  \end{tikzpicture}
  \caption{OSULeaf}
  \end{subfigure}
  \begin{subfigure}[t]{0.48\linewidth}
  \centering
  \begin{tikzpicture}[spy using outlines={circle, magnification=3, size=2cm, connect spies, every spy on node/.append style={thick}}]
    \begin{axis}[
      xmax=7.6, xmin=-0.01,
      name=plot3, 
      y label style={align=center, xshift=-1mm,yshift=\ylabelyshift},
      x label style={yshift=\xlabelyshift},
      xlabel={Wall clock time (\si{\hour})}, 
      ylabel={{Test Accuracy (\si{\percent})}},
      height=\plotheight, width=\plotwidth,
      clip mode=individual,
      legend style={at={(0.99, 0.2)},anchor=east,nodes={scale=0.5, transform shape}}, 
      ]

      \addplot[mark=none, color=blue] table [x=timestamp, y=accuracy, col sep=comma] {images/AccuracyFinalElectricDevices20True.csv};
      \addlegendentry{\textsc{RockNet}};
      \addplot[color=blue, line width=1pt, mark=none, dotted, dash pattern=on 1pt off 1pt, forget plot] coordinates {(-1,67.5) (11,67.5)};

      \addplot[mark=none, color=red] table [x=timestamp, y=accuracy, col sep=comma] {images/AccuracyFinalElectricDevicesNN1True.csv};
      \addlegendentry{AIfES};
      \addplot[color=red, line width=1pt, mark=none, dotted, dash pattern=on 1pt off 1pt, forget plot] coordinates {(-1,51.5) (11,51.5)};

    \end{axis}
  \end{tikzpicture}
  \caption{ElectricDevices}
  \end{subfigure}
  \begin{subfigure}[t]{0.48\linewidth}
  \centering
  \begin{tikzpicture}[spy using outlines={circle, magnification=3, size=2cm, connect spies, every spy on node/.append style={thick}}]
    \begin{axis}[
      xmax=5.01, 
      xmin=-0.01,
      y label style={align=center, xshift=-1mm,yshift=\ylabelyshift},
      x label style={yshift=\xlabelyshift},
      name=plot3, xlabel={Wall clock time (\si{\hour})}, 
      ylabel={{Test Accuracy (\si{\percent})}},
      height=\plotheight, width=\plotwidth,
      clip mode=individual,
      legend style={at={(0.99, 0.2)},anchor=east,nodes={scale=0.5, transform shape}}, 
      ]

      \addplot[mark=none, color=blue] table [x=timestamp, y=accuracy, col sep=comma] {images/AccuracyFinalFaceAll20True.csv};
      \addlegendentry{\textsc{RockNet}}
      \addplot[color=blue, line width=1pt, mark=none, dotted, dash pattern=on 1pt off 1pt, forget plot] coordinates {(-1,72.5) (15,72.5)};

      \addplot[mark=none, color=red] table [x=timestamp, y=accuracy, col sep=comma] {images/AccuracyFinalFaceAllNN1True.csv};
      \addlegendentry{AIfES}
      \addplot[color=red, line width=1pt, mark=none, dotted, dash pattern=on 1pt off 1pt, forget plot] coordinates {(-1,47.5) (15,47.5)};

    \end{axis}
  \end{tikzpicture}
  \caption{FaceAll}
  \end{subfigure}
  \vspace{-0.4cm}
  \caption{\textsc{RockNet} training on three different datasets compared to on-device NN training (AIfES~\cite{wulfert2024aifes}). \capt{Dashed lines show maximum accuracy reached.}}
  \label{fig:RockNet}
  \vspace{-0.5cm}
\end{figure}

Figure~\ref{fig:RockNet} and Table~\ref{tab:hardwareexperiments} compare \textsc{RockNet} running on 20 devices with AIfES running on a single device.
We observe that \textsc{RockNet} significantly outperforms AIfES in accuracy.
For example, on the OSULeaf dataset, \textsc{RockNet} achieves an accuracy of \SI{92.9}{\percent}, whereas AIfES achieves only an accuracy of \SI{39}{\percent}.
At the same time, \textsc{RockNet} needs only \SI{55}{\kilo\byte} of RAM per device, whereas AIfES needs more than three times as much memory.
The higher accuracy of \textsc{RockNet} requires a longer convergence time, ranging between \SI{5}{\hour} and \SI{7.6}{\hour} for the three datasets.
This also entails higher energy costs: \textsc{RockNet} consumes up to $57\times$ more energy per device until it converges and $5\times$ more energy until its accuracy exceeds the accuracy of AIfES.

Nevertheless, despite these costs, \textsc{RockNet} is suitable for scenarios where learning is an infrequent activity.
Consider, for example, the power tool scenario illustrated above.
As long as the factory's processes remain unchanged, relearning the model is unnecessary.
Only when the factory begins producing a new product, the model may become inaccurate and require retraining.
Since product cycles span months to years, \textsc{RockNet} is a viable solution.
Waiting a few more hours for a model with a significantly higher accuracy enables better decision-making and higher resource savings later on.

\subsubsection{Scaling Behavior} 
\label{sec:scaling}

The following experiments take a closer look at \textsc{RockNet}'s performance
on the number of devices.

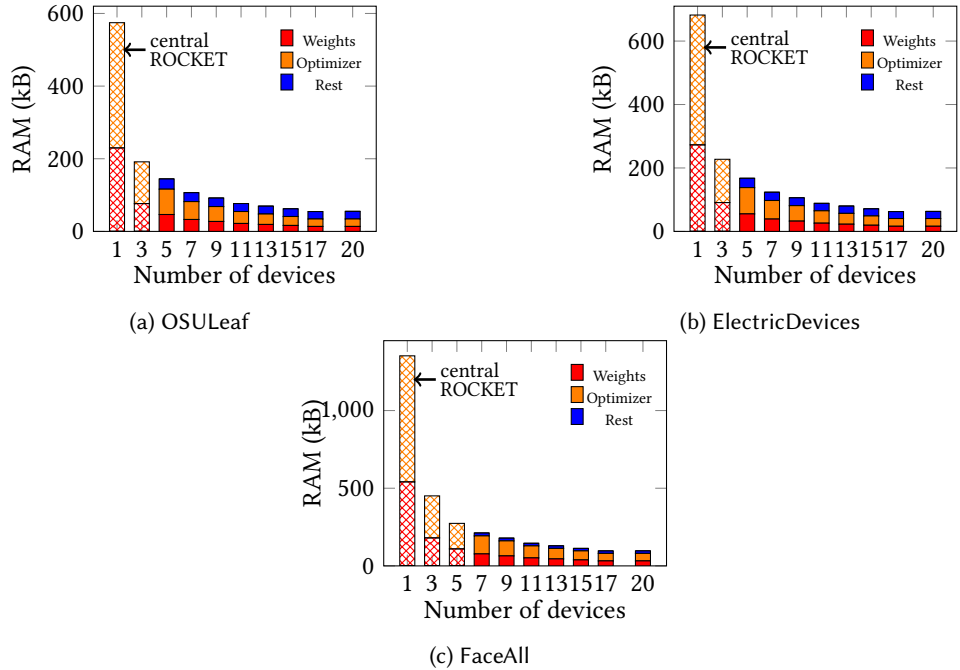
\begin{figure*}[h]
  \newcommand{\ylabelyshiftt}{-3mm}
  \centering
  \begin{subfigure}[t]{0.48\linewidth}
    \centering
    \begin{tikzpicture}[spy using outlines={circle, magnification=3, size=2cm, connect spies, every spy on node/.append style={thick}}]
      \begin{axis}[
        ybar stacked,
        bar width=0.2cm,
        ymax=620, ymin=-0.01,
        xtick={1, 3, 5, 7, 9, 11, 13, 15, 17, 20},
        name=plot3, xlabel={Number of devices}, 
        ylabel={{RAM (\qty{}{\kilo\byte})}},
        y label style={xshift=0mm,yshift=\ylabelyshiftt},
        xlabel style={yshift=2mm},
        height=\plotheightresources, width=\plotwidthresources,
        clip mode=individual,
        legend style={at={(0.99, 0.75)},anchor=east,nodes={scale=0.6, transform shape},fill=none,draw=none}, 
		legend image post style={scale=0.75}, 
        ]

        \addplot[mark=none, color=black, fill=red] coordinates {
          (1,0)
          (3,0)
          (5,46.512)
          (7,32.832)
          (9,27.36)
          (11,21.888)
          (13,19.152)
          (15,16.416)
          (17,13.68)
          (20,13.68)
        };
        \addlegendentry{Weights}

        \addplot[mark=none, color=black, fill=orange] coordinates {
          (1,0)
          (3,0)
          (5,69.768)
          (7,49.248000000000005)
          (9,41.04)
          (11,32.832)
          (13,28.728)
          (15,24.624000000000002)
          (17,20.52)
          (20,20.52)
        };
        \addlegendentry{Optimizer}
        
        \addplot[mark=none, color=black, fill=blue] coordinates {
          (1,0)
          (3,0)
          (5,28.02000000000001)
          (7,24.519999999999982)
          (9,23.30000000000001)
          (11,21.979999999999997)
          (13,21.520000000000003)
          (15,21.059999999999995)
          (17,20.500000000000007)
          (20,20.9)  
        };
        \addlegendentry{Rest}

        \addplot[mark=none, color=black, fill=red, pattern=crosshatch, pattern color=red, forget plot] coordinates {
          (1,229.824)
          (3,76.608)
          (5,0)
          (7,0)
          (9,0)
          (11,0)
          (13,0)
          (15,0)
          (17,0)
          (20,0)
        };

        \addplot[mark=none, color=black, fill=orange, pattern=crosshatch, pattern color=orange, forget plot] coordinates {
          (1,344.736)
          (3,114.912)
          (5,0)
          (7,0)
          (9,0)
          (11,0)
          (13,0)
          (15,0)
          (17,0)
          (20,0)
        };
        \newcommand{\ycoord}{500}
        \coordinate (a) at (axis cs: 3.3, \ycoord);
        \node[yshift=1.3mm,text width=0.5cm] at (axis cs: 5.2,\ycoord) {\fontsize{8pt}{0pt}\selectfont central};
        \node[yshift=-1.3mm,text width=0.5cm] at (axis cs: 5.2,\ycoord) {\fontsize{8pt}{0pt}\selectfont ROCKET};
        \coordinate (b) at (axis cs: 1.5, \ycoord);
      \end{axis}
    \draw[black,line width=1pt,->] (a) -- (b);
    \end{tikzpicture}
    \caption{OSULeaf}
  \end{subfigure}
  \begin{subfigure}[t]{0.48\linewidth}
    \centering
    \begin{tikzpicture}[spy using outlines={circle, magnification=3, size=2cm, connect spies, every spy on node/.append style={thick}}]
      \begin{axis}[
        ybar stacked,
        bar width=0.2cm,
        ymax=710, ymin=-0.01,
        xtick={1, 3, 5, 7, 9, 11, 13, 15, 17, 20},
        name=plot3, xlabel={Number of devices}, 
        ylabel={{RAM (\qty{}{\kilo\byte})}},
        y label style={xshift=0mm,yshift=\ylabelyshiftt},
        xlabel style={yshift=2mm},
        height=\plotheightresources, width=\plotwidthresources,
        clip mode=individual,
        legend style={at={(0.99, 0.75)},anchor=east,nodes={scale=0.6, transform shape},fill=none,draw=none}, 
		legend image post style={scale=0.75}, 
        ]

        \addplot[mark=none, color=black, fill=red] coordinates {
          (1,0)
          (3,0)
          (5,55.216)
          (7,38.976)
          (9,32.48)
          (11,25.984)
          (13,22.736)
          (15,19.488)
          (17,16.24)
          (20,16.24)
        };
        \addlegendentry{Weights}

        \addplot[mark=none, color=black, fill=orange] coordinates {
          (1,0)
          (3,0)
          (5,82.82400000000001)
          (7,58.464)
          (9,48.71999999999999)
          (11,38.976000000000006)
          (13,34.104)
          (15,29.232)
          (17,24.359999999999996)
          (20,24.359999999999996)
        };
        \addlegendentry{Optimizer}
        
        \addplot[mark=none, color=black, fill=blue] coordinates {
          (1,0)
          (3,0)
          (5,29.26000000000002)
          (7,25.860000000000014)
          (9,24.599999999999994)
          (11,23.33999999999999)
          (13,22.86)
          (15,22.380000000000003)
          (17,21.9)
          (20,22.3)
        };
        \addlegendentry{Rest}

        \addplot[mark=none, color=black, fill=red, pattern=crosshatch, pattern color=red, forget plot] coordinates {
          (1,272.832)
          (3,90.944)
          (5,0)
          (7,0)
          (9,0)
          (11,0)
          (13,0)
          (15,0)
          (17,0)
          (20,0)
        };

        \addplot[mark=none, color=black, fill=orange, pattern=crosshatch, pattern color=orange, forget plot] coordinates {
          (1,409.248)
          (3,136.416)
          (5,0)
          (7,0)
          (9,0)
          (11,0)
          (13,0)
          (15,0)
          (17,0)
          (20,0)
        };

        \newcommand{\ycoord}{580}
        \coordinate (a) at (axis cs: 3.3, \ycoord);
        \node[yshift=1.3mm,text width=0.5cm] at (axis cs: 5.2,\ycoord) {\fontsize{8pt}{0pt}\selectfont central};
        \node[yshift=-1.3mm,text width=0.5cm] at (axis cs: 5.2,\ycoord) {\fontsize{8pt}{0pt}\selectfont ROCKET};
        \coordinate (b) at (axis cs: 1.5, \ycoord);
      \end{axis}
    \draw[black,line width=1pt,->] (a) -- (b);
    \end{tikzpicture}
    \caption{ElectricDevices}
  \end{subfigure}
  \begin{subfigure}[t]{0.48\linewidth}
    \centering
    \begin{tikzpicture}[spy using outlines={circle, magnification=3, size=2cm, connect spies, every spy on node/.append style={thick}}]
      \begin{axis}[
        ybar stacked,
        bar width=0.2cm,
        ymax=1450, ymin=-0.01,
        xtick={1, 3, 5, 7, 9, 11, 13, 15, 17, 20},
        name=plot3, xlabel={Number of devices}, 
        ylabel={{RAM (\qty{}{\kilo\byte})}},
        y label style={xshift=0mm,yshift=\ylabelyshiftt},
        xlabel style={yshift=2mm},
        height=\plotheightresources, width=\plotwidthresources,
        clip mode=individual,
        legend style={at={(0.99, 0.75)},anchor=east,nodes={scale=0.6, transform shape},fill=none,draw=none}, 
		    legend image post style={scale=0.75}, 
        ]

        \addplot[mark=none, color=black, fill=red] coordinates {
          (1,0)
          (3,0)
          (5,0)
          (7,77.28)
          (9,64.39999999999999)
          (11,51.52)
          (13,45.080000000000005)
          (15,38.64)
          (17,32.199999999999996)
          (20,32.199999999999996)
        };
        \addlegendentry{Weights}

        \addplot[mark=none, color=black, fill=orange] coordinates {
          (1,0)
          (3,0)
          (5,0)
          (7,115.91999999999999)
          (9,96.6)
          (11,77.28)
          (13,67.62)
          (15,57.959999999999994)
          (17,48.3)
          (20,48.3)
        };
        \addlegendentry{Optimizer}
        
        \addplot[mark=none, color=black, fill=blue] coordinates {
          (1,0)
          (3,0)
          (5,0)
          (7,18.200000000000017)
          (9,17.30000000000001)
          (11,16.399999999999977)
          (13,16.099999999999994)
          (15,15.800000000000011)
          (17,15.5)
          (20,15.8)
        };
        \addlegendentry{Rest}

        \addplot[mark=none, color=black, fill=red, pattern=crosshatch, pattern color=red, forget plot] coordinates {
          (1,540.96)
          (3,180.32000000000002)
          (5,109.48)
          (7,0)
          (9,0)
          (11,0)
          (13,0)
          (15,0)
          (17,0)
          (20,0)
        };

        \addplot[mark=none, color=black, fill=orange, pattern=crosshatch, pattern color=orange, forget plot] coordinates {
          (1,811.44)
          (3,270.48)
          (5,164.22)
          (7,0)
          (9,0)
          (11,0)
          (13,0)
          (15,0)
          (17,0)
          (20,0)
        };

        \newcommand{\ycoord}{1200}
        \coordinate (a) at (axis cs: 3.3, \ycoord);
        \node[yshift=1.3mm,text width=0.5cm] at (axis cs: 5.2,\ycoord) {\fontsize{8pt}{0pt}\selectfont central};
        \node[yshift=-1.3mm,text width=0.5cm] at (axis cs: 5.2,\ycoord) {\fontsize{8pt}{0pt}\selectfont ROCKET};
        \coordinate (b) at (axis cs: 1.5, \ycoord);
      \end{axis}
    \draw[black,line width=1pt,->] (a) -- (b);
    \end{tikzpicture}
    \caption{FaceAll}
  \end{subfigure}
  \caption{Per device memory consumption of \textsc{RockNet} regarding network size. \textit{
    Crosshatched bars show predictions for configurations that would require more than \SI{256}{KB} of available RAM per device, thus exceeding the hardware limits. 
    The first bar (1 device) corresponds to central ROCKET.
    It significantly exceeds the available memory.
    Split learning via \textsc{RockNet} enables ROCKET on this hardware, with the memory consumption falling with the number of devices.}}
  \label{fig:memory}
\end{figure*}

\fakepar{Memory (Figure~\ref{fig:memory})}
Central ROCKET consumes up to \SI{1352}{\kilo\byte} of RAM, which exceeds the available \SI{256}{\kilo\byte} per device by \SI{428}{\percent}. 
Therefore, training and inferencing MiniROCKET on a single device is not feasible. 
\textsc{RockNet}'s distributed execution enables ROCKET for ultra-low-power hardware. Depending on the number of classes, at least 5 to 7 devices are required for MiniROCKET.

The memory consumption falls approximately inversely proportional with the number of devices ($\sim 1 / N$) as derived in Section~\ref{sec:analysis_res_consumption:memory}.
Initially, there is a significant reduction in memory usage with a low number of devices, which then becomes more gradual as the number of devices increases.
For example, when running \textsc{RockNet} on five devices for ElectricDevices, the per-device memory consumption is reduced by \SI{75.5}{\percent} compared to central ROCKET. 
As a result, \textsc{RockNet} on five devices utilizes only \SI{65.4}{\percent} of the available memory. 
Further, scaling up to 20 devices reduces memory consumption to just \SI{24.6}{\percent} of the available memory, which is a relative reduction of \SI{90}{\percent} compared to central ROCKET.
The largest memory consumer is the ADAM optimizer (\SI{66}{\percent} for \SI{32}{\bit} and \SI{48}{\percent} for \SI{8}{\bit} ADAM).
The weights themselves are the second-largest memory consumers (\SI{32}{\percent}), followed by the features and other program code (\SI{20}{\percent}).

Since each device only has \SI{256}{\kilo\byte} of RAM, training and inferencing MiniROCKET on a single device, i.e., central ROCKET, is not feasible.
Depending on the number of classes, at least 5 to 7 devices are required for MiniROCKET.
Thus, \textsc{RockNet}'s distributed execution enables ROCKET for ultra-low-power hardware.

As \textsc{RockNet}'s memory consumption falls with the number of devices, we can reduce per-device memory by including more devices in \textsc{RockNet}. 
This allows us to adapt the memory consumption to the devices' available memory budget, e.g., when parts of the memory are occupied by other applications.

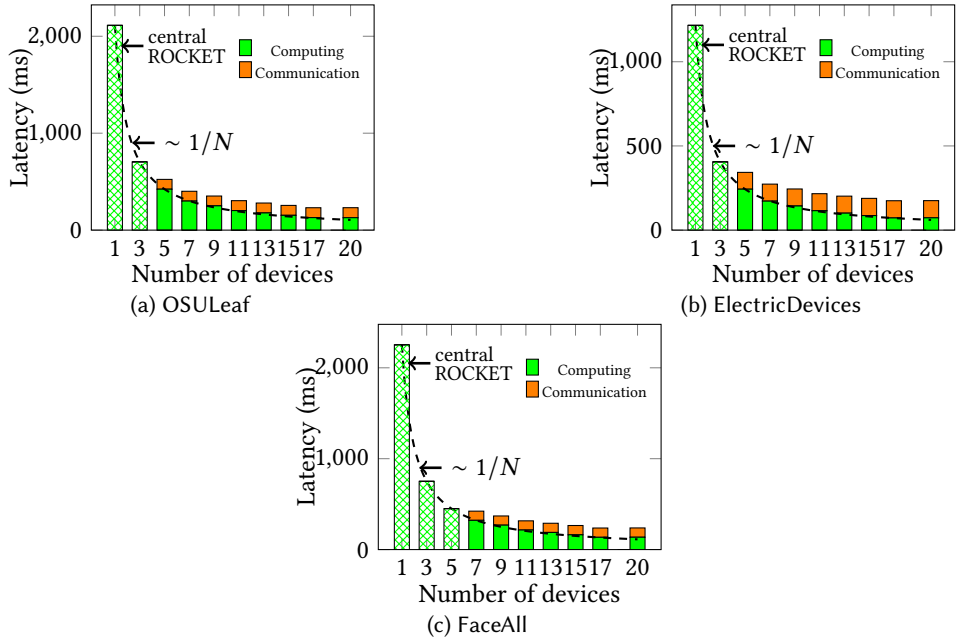
\begin{figure*}[h]
  \newcommand{\ylabelyshiftt}{-3mm}
  \centering
  \begin{subfigure}[t]{0.48\linewidth}
    \centering
    \begin{tikzpicture}[spy using outlines={circle, magnification=3, size=2cm, connect spies, every spy on node/.append style={thick}}]
      \begin{axis}[
        ybar stacked,
        bar width=0.2cm,
        ymin=-0.01,
        xtick={1, 3, 5, 7, 9, 11, 13, 15, 17, 20},
        name=plot3, xlabel={Number of devices}, 
        ylabel={{Latency (\SI{}{\milli\second})}},
        ylabel style={yshift=\ylabelyshiftt},
        xlabel style={yshift=2mm},
        height=\plotheightresources, width=\plotwidthresources,
        clip mode=individual,
        legend style={at={(0.99, 0.75)},anchor=east,nodes={scale=0.6, transform shape},fill=none,draw=none}, 
		legend image post style={scale=0.75}, 
        ]

        \addplot[mark=none, color=black, fill=green, pattern=crosshatch, pattern color=green, forget plot] coordinates {
          (1,2112.31)
          (3,704.1033333333334)
          (5,0)
          (7,0)
          (9,0)
          (11,0)
          (13,0)
          (15,0)
          (17,0)
          (19,0)
        };

        \addplot[mark=none, color=black, thick, dashed, domain=1:20, samples=100, sharp plot, stack plots=false, forget plot] {2112.31/x};

        \addplot[mark=none, color=black, fill=green] coordinates {
          (1,0)
          (3,0)
          (5,422.462)
          (7,299.842)
          (9,250.814)
          (11,201.736)
          (13,177.374)
          (15,152.869)
          (17,128.893)
          (20,128.07)
        };
        \addlegendentry{Computing}

        \addplot[mark=none, color=black, fill=orange] coordinates {
          (1,0)
          (3,0)
          (5,100.95)
          (7,100.95)
          (9,100.95)
          (11,102.15)
          (13,102.15)
          (15,102.15)
          (17,102.15)
          (20,103.35)
        };
        \addlegendentry{Communication}
        \newcommand{\ycoord}{1900}
        \coordinate (a) at (axis cs: 3.3, \ycoord);
        \node[yshift=1.3mm,text width=0.5cm] at (axis cs: 5.2,\ycoord) {\fontsize{8pt}{0pt}\selectfont central};
        \node[yshift=-1.3mm,text width=0.5cm] at (axis cs: 5.2,\ycoord) {\fontsize{8pt}{0pt}\selectfont ROCKET};
        \coordinate (b) at (axis cs: 1.5, \ycoord);

        \newcommand{\ycoordtwo}{900}
        \coordinate (c) at (axis cs: 4.2, \ycoordtwo);
        \node[text width=1.0cm] at (axis cs: 8,\ycoordtwo) {$\sim1/N$};
        \coordinate (d) at (axis cs: 2.4, \ycoordtwo);
      \end{axis}
    \draw[black,line width=1pt,->] (a) -- (b);
    \draw[black,line width=1pt,->] (c) -- (d);
    \end{tikzpicture}
    \vspace{-0.2cm}
    \caption{OSULeaf}
  \end{subfigure}
  \begin{subfigure}[t]{0.48\linewidth}
    \centering
    \begin{tikzpicture}[spy using outlines={circle, magnification=3, size=2cm, connect spies, every spy on node/.append style={thick}}]
      \begin{axis}[
        ybar stacked,
        bar width=0.2cm,
        ymin=-0.01,
        xtick={1, 3, 5, 7, 9, 11, 13, 15, 17, 20},
        name=plot3, xlabel={Number of devices}, 
        ylabel={{Latency (\SI{}{\milli\second})}},
        ylabel style={yshift=\ylabelyshiftt},
        xlabel style={yshift=2mm},
        height=\plotheightresources, width=\plotwidthresources,
        clip mode=individual,
        legend style={at={(0.99, 0.75)},anchor=east,nodes={scale=0.6, transform shape},fill=none,draw=none}, 
		legend image post style={scale=0.75}, 
        ]

        \addplot[mark=none, color=black, fill=green, pattern=crosshatch, pattern color=green, forget plot] coordinates {
          (1,1215.9850000000001)
          (3,405.3283333333334)
          (5,0)
          (7,0)
          (9,0)
          (11,0)
          (13,0)
          (15,0)
          (17,0)
          (19,0)
        };
        \addplot[mark=none, color=black, thick, dashed, domain=1:20, samples=100, sharp plot, stack plots=false, forget plot] {1215.9850000000001/x};

        \addplot[mark=none, color=black, fill=green] coordinates {
          (1,0)
          (3,0)
          (5,243.197)
          (7,172.242)
          (9,143.445)
          (11,115.428)
          (13,100.883)
          (15,86.482)
          (17,72.385)
          (20,72.991)
        };
        \addlegendentry{Computing}

        \addplot[mark=none, color=black, fill=orange] coordinates {
          (1,0)
          (3,0)  
          (5,99.75)
          (7,100.95)
          (9,100.95)
          (11,100.95)
          (13,100.95)
          (15,102.15)
          (17,102.15)
          (20,102.15)
        };
        \addlegendentry{Communication}
        \newcommand{\ycoord}{1100}
        \coordinate (a) at (axis cs: 3.3, \ycoord);
        \node[yshift=1.3mm,text width=0.5cm] at (axis cs: 5.2,\ycoord) {\fontsize{8pt}{0pt}\selectfont central};
        \node[yshift=-1.3mm,text width=0.5cm] at (axis cs: 5.2,\ycoord) {\fontsize{8pt}{0pt}\selectfont ROCKET};
        \coordinate (b) at (axis cs: 1.5, \ycoord);

        \newcommand{\ycoordtwo}{500}
        \coordinate (c) at (axis cs: 4.2, \ycoordtwo);
        \node[text width=1.0cm] at (axis cs: 8,\ycoordtwo) {$\sim1/N$};
        \coordinate (d) at (axis cs: 2.4, \ycoordtwo);
      \end{axis}
    \draw[black,line width=1pt,->] (a) -- (b);
    \draw[black,line width=1pt,->] (c) -- (d);
    \end{tikzpicture}
    \vspace{-0.2cm}
    \caption{ElectricDevices}
  \end{subfigure}
  \begin{subfigure}[t]{0.48\linewidth}
    \centering
    \begin{tikzpicture}[spy using outlines={circle, magnification=3, size=2cm, connect spies, every spy on node/.append style={thick}}]
      \begin{axis}[
        ybar stacked,
        bar width=0.2cm,
        ymin=-0.01,
        xtick={1, 3, 5, 7, 9, 11, 13, 15, 17, 20},
        name=plot3, xlabel={Number of devices}, 
        ylabel={{Latency (\SI{}{\milli\second})}},
        ylabel style={yshift=\ylabelyshiftt},
        xlabel style={yshift=2mm},
        height=\plotheightresources, width=\plotwidthresources,
        clip mode=individual,
        legend style={at={(0.99, 0.75)},anchor=east,nodes={scale=0.6, transform shape},fill=none,draw=none}, 
		legend image post style={scale=0.75}, 
        ]

        \addplot[mark=none, color=black, fill=green, pattern=crosshatch, pattern color=green, forget plot] coordinates {
          (1,2252.831)
          (3,750.9436666666667)
          (5,450.56620000000004)
          (7, 0)
          (9,0)
          (11,0)
          (13,0)
          (15,0)
          (17,0)
          (19,0)
        };

        \addplot[mark=none, color=black, thick, dashed, domain=1:20, samples=100, sharp plot, stack plots=false, forget plot] {2252.831/x};
        
        \addplot[mark=none, color=black, fill=green] coordinates {
          (1,0)
          (3,0)
          (5,0)
          (7,321.833)
          (9,268.916)
          (11,215.24)
          (13,188.672)
          (15,162.736)
          (17,135.141)
          (20,136)
        };
        \addlegendentry{Computing}

        \addplot[mark=none, color=black, fill=orange] coordinates {
          (1,0)
          (3,0)
          (5,0)
          (7,100.95)
          (9,100.95)
          (11,100.95)
          (13,100.95)
          (15,102.15)
          (17,102.15)
          (20,102.15)
        };
        \addlegendentry{Communication}

        \newcommand{\ycoord}{2050}
        \coordinate (a) at (axis cs: 3.3, \ycoord);
        \node[yshift=1.3mm,text width=0.5cm] at (axis cs: 5.2,\ycoord) {\fontsize{8pt}{0pt}\selectfont central};
        \node[yshift=-1.3mm,text width=0.5cm] at (axis cs: 5.2,\ycoord) {\fontsize{8pt}{0pt}\selectfont ROCKET};
        \coordinate (b) at (axis cs: 1.5, \ycoord);

                \newcommand{\ycoordtwo}{900}
        \coordinate (c) at (axis cs: 4.2, \ycoordtwo);
        \node[text width=1.0cm] at (axis cs: 8,\ycoordtwo) {$\sim1/N$};
        \coordinate (d) at (axis cs: 2.4, \ycoordtwo);
      \end{axis}
    \draw[black,line width=1pt,->] (a) -- (b);
    \draw[black,line width=1pt,->] (c) -- (d);
    \end{tikzpicture}
    \vspace{-0.2cm}
    \caption{FaceAll}
  \end{subfigure}
  \vspace{-0.4cm}
  \caption{Latency of \textsc{RockNet} per training step. 
  \textit{
  The latency for computing falls inversely proportional with number of devices, while the communication's latency only grows slightly.
  As a result, the latency first falls drastically for a low number of devices and then more gradually for a higher number.}
    }
  \label{fig:latency}
\end{figure*}

\fakepar{Latency (Figure~\ref{fig:latency})}
We measured the maximum latency of \textsc{RockNet} per step as the number of devices increases measured over 500 training and 500 evaluation steps.
The latency first decreases drastically for a low number of devices and only slightly for a higher number of devices.
For FaceAll, for example, computing latency scales from 5 to 15 devices with a factor of around $2.76$ and to 20 devices with a factor of $3.3$. The gaps to ``perfect'' scaling (i.e., factor 3 and 4) are caused by overhead every device has to do, like processing, preparing communication messages and the calculation of softmax scores.
In absolute values, scaling from 5 to 15 devices reduces computing latency by \SI{269.59}{\milli\second} and by \SI{294.39}{\milli\second} for 20 devices.
On the other side, communication latency stays almost constant at \SI{100}{\milli\second}, with an increase of only \SI{3}{\percent} or \SI{3}{\milli\second} from 5 to 20 devices.
Consequently, when scaling from five to 20 devices, the overall latency reduces by \SI{55}{\percent}.

Especially compared to central ROCKET, \textsc{RockNet} significantly decreases latency.
For instance, in our complete training run for ElectricDevices discussed in Section~\ref{sec:aifesvsrocknet}, central ROCKET would take more than two days to converge. 
In contrast, \textsc{RockNet} with five devices speeds up training by
$2.3\times$, reducing it to \SI{22}{\hour}. Finally, using 20 devices with \textsc{RockNet}, convergence is achieved in just \SI{7.5}{\hour}, representing a 
$6.9\times$ speedup.  
This improvement stems primarily from a significant reduction in computing latency, which dropped from \SI{1215.99}{\milli\second} to \SI{72.99}{\milli\second} ($16.6\times$) compared to central ROCKET.
This reduction is counteracted by additional \SI{100}{\milli\second} of communication overhead, which amounts to roughly \SI{58}{\percent} of the total latency for 20 devices.
However, as it is significantly lower than the speedup in computing, \textsc{RockNet} still achieves an overall speedup of $6.9\times$.

The reason for this behavior is that the compute time decreases almost inversely ($\sim 1/N$) with the number of devices (cf. Section~\ref{sec:analysis_res_consumption:flops}). 
Therefore, it decreases significantly with a low number of devices and only slightly with a high number. 
On the other hand, communication time increases only slightly due to Mixer’s order-optimal scaling behavior~\cite{Mixer}.
Thus, \textsc{RockNet} efficiently pools compute power across multiple devices, significantly speeding up the training process as more devices are added.

\begin{figure*}[h]
  \newcommand{\ylabelyshiftt}{-4mm}
  \newcommand{\errorbargreen}{green!50!black}
  \newcommand{\errorbarred}{red!50!black}
  \newcommand{\errorbarorange}{orange!50!black}
  \centering
  \begin{subfigure}[t]{0.48\linewidth}
    \centering
    \begin{tikzpicture}[spy using outlines={circle, magnification=3, size=2cm, connect spies, every spy on node/.append style={thick}}]
      \begin{axis}[
        ybar stacked,
        bar width=0.2cm,
        ymin=-0.01,
        xtick={1, 3, 5, 7, 9, 11, 13, 15, 17, 20},
        name=plot3, xlabel={Number of devices}, 
        ylabel={{Energy (\SI{}{\milli\joule})}},
        ylabel style={align=center,yshift=\ylabelyshiftt},
        xlabel style={yshift=2mm},
        height=\plotheightresources, width=\plotwidthresources,
        clip mode=individual,
        legend style={at={(0.99, 0.75)},anchor=east,nodes={scale=0.6, transform shape},fill=none,draw=none}, 
		legend image post style={scale=0.75}, 
        ]
        
        \addplot[mark=none, color=black, fill=green, pattern=crosshatch, pattern color=green, forget plot] coordinates {
          (1,15.698932823625002)
          (3,5.232977607875001)
          (5,0)
          (7,0)
          (9,0)
          (11,0)
          (13,0)
          (15,0)
          (17,0)
          (19,0)
        };

        \addplot[mark=none, color=black, thick, dashed, domain=1:20, samples=100, sharp plot, stack plots=false, forget plot] {15.698932823625002/x};

        \addplot[
          color=black, fill=green,
            error bars/.cd,
                y dir=both,
                error bar style={color=\errorbargreen},
                y explicit,
        ] coordinates {
          (1,0)
          (3,0)  
          (5,3.1397865647250005)
(7,2.234686077285)
(9,1.8700164294600001)
(11,1.51115708259)
(13,1.3316294551349999)
(15,1.1499838883250002)
(17,0.9706427793000001)
(20,0.957577360575)
        }; 
        \addlegendentry{Computing}

        \addplot[mark=none, color=black, fill=orange,error bars/.cd,
        y dir=both,
        error bar style={color=\errorbarorange},
        y explicit] coordinates {
          (1,0)
          (3,0)  
          (5,0.5324714356429412)
(7,0.703868598020294)
(9,0.8156055677673528)
(11,0.8941562462481818)
(13,0.9636255016459091)
(15,1.0662679830695456)
(17,1.1636474611186363)
(20,1.1801635912053658)
        };
        \addlegendentry{Communication}
        \newcommand{\ycoord}{14.1}
        \coordinate (a) at (axis cs: 3.3, \ycoord);
        \node[yshift=1.3mm,text width=0.5cm] at (axis cs: 5.2,\ycoord) {\fontsize{8pt}{0pt}\selectfont central};
        \node[yshift=-1.3mm,text width=0.5cm] at (axis cs: 5.2,\ycoord) {\fontsize{8pt}{0pt}\selectfont ROCKET};
        \coordinate (b) at (axis cs: 1.5, \ycoord);
      
        \newcommand{\ycoordtwo}{7} 
        \coordinate (c) at (axis cs: 4.2, \ycoordtwo);
        \node[text width=1.0cm] at (axis cs: 8,\ycoordtwo) {$\sim1/N$};
        \coordinate (d) at (axis cs: 2.4, \ycoordtwo);
      \end{axis}
    \draw[black,line width=1pt,->] (a) -- (b);
    \draw[black,line width=1pt,->] (c) -- (d);
    \end{tikzpicture}
    \vspace{-0.2cm}
    \caption{OSULeaf}
  \end{subfigure}
  \begin{subfigure}[t]{0.48\linewidth}
    \centering
    \begin{tikzpicture}[spy using outlines={circle, magnification=3, size=2cm, connect spies, every spy on node/.append style={thick}}]
      \begin{axis}[
        ybar stacked,
        bar width=0.2cm,
        ymin=-0.01,
        xtick={1, 3, 5, 7, 9, 11, 13, 15, 17, 20},
        name=plot3, xlabel={Number of devices}, 
        ylabel={{Energy (\SI{}{\milli\joule})}},
        ylabel style={align=center, yshift=\ylabelyshiftt},
        xlabel style={yshift=2mm},
        height=\plotheightresources, width=\plotwidthresources,
        clip mode=individual,
        legend style={at={(0.99, 0.75)},anchor=east,nodes={scale=0.6, transform shape},fill=none,draw=none}, 
		legend image post style={scale=0.75}, 
        ]

        \addplot[mark=none, color=black, fill=green, pattern=crosshatch, pattern color=green, forget plot] coordinates {
          (1,4.34559806565)
          (3,1.44853268855)
          (5,0)
          (7,0)
          (9,0)
          (11,0)
          (13,0)
          (15,0)
          (17,0)
          (19,0)
        };

        \addplot[mark=none, color=black, thick, dashed, domain=1:20, samples=100, sharp plot, stack plots=false, forget plot] {4.34559806565/x};
        
        \addplot[
          color=black, fill=green,
            error bars/.cd,
                y dir=both,
                error bar style={color=\errorbargreen},
                y explicit,
        ] coordinates {
          (1,0)
          (3,0)  
          (5,0.8691196131300001)
          (7,0.618121724835)
          (9,0.51604687527)
          (11,0.415052973405)
          (13,0.36455883792)
          (15,0.3134801447100001)
          (17,0.263623182135)
          (20,0.26408261037)
        };
        \addlegendentry{Computing}

        \addplot[mark=none, color=black, fill=orange,error bars/.cd,
        y dir=both,
        error bar style={color=\errorbarorange},
        y explicit] coordinates {
          (1,0)
          (3,0)  
          (5,0.4181764584711538)
          (7,0.5259697322717647)
          (9,0.6541818540397057)
          (11,0.7769366723647058)
          (13,0.8670653480214705)
          (15,0.949644636994091)
          (17,1.038169197025909)
          (20,1.2005441710977272)
        };
        \addlegendentry{Communication}
        \newcommand{\ycoord}{3.9}
        \coordinate (a) at (axis cs: 3.3, \ycoord);
        \node[yshift=1.3mm,text width=0.5cm] at (axis cs: 5.2,\ycoord) {\fontsize{8pt}{0pt}\selectfont central};
        \node[yshift=-1.3mm,text width=0.5cm] at (axis cs: 5.2,\ycoord) {\fontsize{8pt}{0pt}\selectfont ROCKET};
        \coordinate (b) at (axis cs: 1.5, \ycoord);

        \newcommand{\ycoordtwo}{2}
        \coordinate (c) at (axis cs: 4.2, \ycoordtwo);
        \node[text width=1.0cm] at (axis cs: 8,\ycoordtwo) {$\sim1/N$};
        \coordinate (d) at (axis cs: 2.4, \ycoordtwo);
      \end{axis}
    \draw[black,line width=1pt,->] (a) -- (b);
    \draw[black,line width=1pt,->] (c) -- (d);
    \end{tikzpicture}
    \vspace{-0.2cm}
    \caption{ElectricDevices}
  \end{subfigure}
  \begin{subfigure}[t]{0.48\linewidth}
    \centering
    \begin{tikzpicture}[spy using outlines={circle, magnification=3, size=2cm, connect spies, every spy on node/.append style={thick}}]
      \begin{axis}[
        ybar stacked,
        bar width=0.2cm,
        ymin=-0.01,
        xtick={1, 3, 5, 7, 9, 11, 13, 15, 17, 20},
        name=plot3, xlabel={Number of devices}, 
        ylabel={{Energy (\SI{}{\milli\joule})}},
        ylabel style={align=center, yshift=\ylabelyshiftt},
        xlabel style={yshift=2mm},
        height=\plotheightresources, width=\plotwidthresources,
        clip mode=individual,
        legend style={at={(0.99, 0.75)},anchor=east,nodes={scale=0.6, transform shape},fill=none,draw=none}, 
		legend image post style={scale=0.75}, 
        ]

        \addplot[mark=none, color=black, fill=green, pattern=crosshatch, pattern color=green, forget plot] coordinates {
          (1,6.960634735845002)
          (3,2.3202115786150004)
          (5,1.3921269471690003)
          (7,0)
          (9,0)
          (11,0)
          (13,0)
          (15,0)
          (17,0)
          (20,0)
        };

        \addplot[mark=none, color=black, thick, dashed, domain=1:20, samples=100, sharp plot, stack plots=false, forget plot] {6.960634735845002/x};

        \addplot[mark=none, color=black, fill=green, 
        error bars/.cd, y dir=both,
        error bar style={color=\errorbargreen},
        y explicit] coordinates {
          (1,0)
          (3,0)  
          (5,0)
          (7,0.9943763908350002)
        (9,0.8349113806950001)
        (11,0.6682824874200001)
        (13,0.5890382629650002)
        (15,0.5079124984650001)
        (17,0.42393124498500007)
          (20,0.42234265429500006)
        };
        \addlegendentry{Computing}

        \addplot[mark=none, color=black, fill=orange, 
        error bars/.cd,
        y dir=both,
        error bar style={color=\errorbarorange},
        y explicit] coordinates {
          (1,0)
          (3,0)  
          (5,0)
          (7,0.5604762720723528)
          (9,0.660609375587647)
          (11,0.8003752624914704)
          (13,0.8505714422549999)
          (15,0.9386322048095455)
          (17,1.0196793792327272)
          (20,1.2358485736336364)
        };
        \addlegendentry{Communication}
        \newcommand{\ycoord}{6.3}
        \coordinate (a) at (axis cs: 3.3, \ycoord);
        \node[yshift=1.3mm,text width=0.5cm] at (axis cs: 5.2,\ycoord) {\fontsize{8pt}{0pt}\selectfont central};
        \node[yshift=-1.3mm,text width=0.5cm] at (axis cs: 5.2,\ycoord) {\fontsize{8pt}{0pt}\selectfont ROCKET};
        \coordinate (b) at (axis cs: 1.5, \ycoord);

        \newcommand{\ycoordtwo}{3}
        \coordinate (c) at (axis cs: 4.2, \ycoordtwo);
        \node[text width=1.0cm] at (axis cs: 8,\ycoordtwo) {$\sim1/N$};
        \coordinate (d) at (axis cs: 2.4, \ycoordtwo);
      \end{axis}
    \draw[black,line width=1pt,->] (a) -- (b);
    \draw[black,line width=1pt,->] (c) -- (d);
    \end{tikzpicture}
    \vspace{-0.2cm}
    \caption{FaceAll}
  \end{subfigure}
  \vspace{-0.4cm}
  \caption{Energy consumption of \textsc{RockNet} per device and per training step. 
  \textit{We measured mean energy consumption, the most important measure for battery powered devices, for 500 training and 500 inference steps.
  The energy for computing falls inversely proportional with number of devices, while the communication's energy consumption grows linearly.
  As a result, for a specific number of devices, the energy consumption is minimal.
  }}
  \label{fig:energy}
  \vspace{-4mm}
\end{figure*}
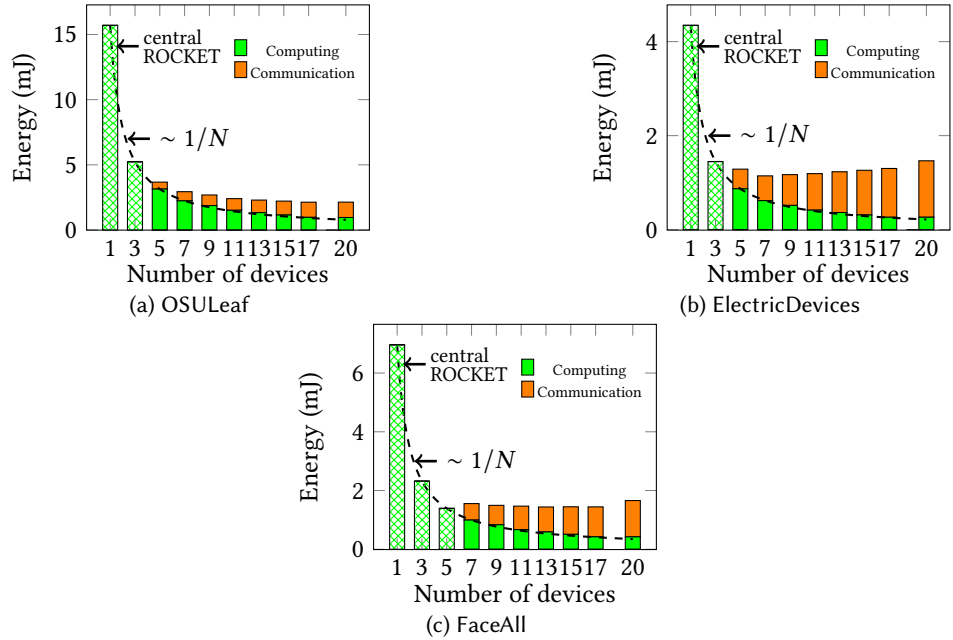

\fakepar{Energy (Figure~\ref{fig:energy})}
Figure~\ref{fig:energy} shows the mean energy consumption per training step as the device number increases measured over 500 training and 500 evaluation steps.
We measured mean consumption as it is the most relevant one for e.g., battery powered devices.

We can see that initially the energy falls drastically at the beginning, reaches a minimum, and then grows slightly.
For OSULeaf, the minimum is reached at 20 devices, for ElectricDevices at 7 and for FaceAll at 15.
The energy required for compute decreases roughly in inverse proportion to the number of devices ($\sim 1/N$), aside from a small overhead (see latency). For instance, on the ElectricDevices dataset, scaling from 5 to 15 devices reduces per-device energy consumption by about $2.77\times$ and from 5 to 20 devices by $3.29\times$.
At the same time, total energy consumption per round follows an affine trend, rising by around \SI{0.05}{\joule} per device, from \SI{0.42}{\joule} for 5 devices to \SI{1.2}{\joule} for 20 devices (a factor of $2.86\times$).
At 20 devices, communication becomes the main contributor, accounting for \SI{81.96}{\percent} of the total energy consumption.

Compared to central ROCKET, \textsc{RockNet} significantly reduces per-device energy consumption.
For example, in the case of ElectricDevices, mean energy consumption falls by \SI{73.6}{\percent} when moving from central ROCKET to \textsc{RockNet} to the minimum at seven devices. 
However, as the number of devices increases to 20, energy consumption grows by \SI{28}{\percent}, leading to a reduction in energy consumption of \SI{66.2}{\percent}.
This behavior is caused by energy for compute decreasing approximately inversely with the number of devices ($\sim 1/N$), while Mixer's energy consumption scales approximately linearly with the number of devices (cf. Section~\ref{sec:analysis_res_consumption}).

\section{Discussion}
\label{sec:discussionbaselines}
Our evaluation demonstrates that \textsc{RockNet} effectively pools the resources of multiple devices to perform learning.
Utilizing more devices reduces the compute load and energy consumption on each device.

Our experiments compared \textsc{RockNet} to two central ML approaches: AIfES and central ROCKET. 
\textsc{RockNet} achieves a significantly higher accuracy than AIfES.
Conversely, AIfES consumes fewer overall resources, particularly in terms of energy and latency. 
However, as discussed in Section~\ref{sec:aifesvsrocknet}, the increased energy consumption and latency of \textsc{RockNet} is justified in scenarios where a model is learned once and then used for a longer time.

\textsc{RockNet} significantly reduces latency, memory usage and per-device energy compared to the execution of ROCKET on a single device (assuming the device is equipped with enough RAM),
As more devices are added, these per-device resource consumption is reduced until a limit is reached.
In our BLE-based hardware implementation, we reach this point for latency at approximately 20 devices, where we achieve a maximum reduction of \SI{89}{\percent} relative to a single-device ROCKET execution for the FaceAll dataset.
Beyond 20 devices, latency begins to rise. 
Regarding energy consumption, this point is reached earlier, between 7 and 15 devices.
This behavior is expected when dividing a fixed-size computing task. While the per-device workload decreases proportional to $1/N$, the communication overhead increases linearly (cf. Section~\ref{sec:analysis_res_consumption}), resulting in diminishing absolute gains, and at some point, in an increase in resource consumption.
The minimum point of energy consumption is reached much earlier than for latency, as power drawn from our chip during receiving/transmitting is around $3.67\times$ higher compared to computing.

Another noteworthy observation is that, while communication latency via Mixer remains nearly constant, energy consumption continues to rise. This occurs because Mixer automatically places devices in energy-saving mode whenever possible. Consequently, sending smaller amounts of data increases the share of power-saving slots relative to RX/TX slots. Conversely, transferring larger amounts of data leaves fewer opportunities for energy-saving intervals, thus increasing overall energy consumption.

We demonstrated in our experiments that split learning across multiple low-power devices is feasible. 
In this way, we avoid the need of a central device. 
In other application (see Sec.~\ref{sec:introduction:contributions}), a central device equipped with more powerful hardware might be preferred. 
Typically, stronger hardware is more energy-efficient (more floating point operations per Watts) and faster. 
In practice, one must evaluate whether using such a device is economically feasible--essentially determining whether the benefits of faster and more energy-efficient training justify the cost of the stronger hardware.

\section{Limitations}
This section addresses the limitations of our work and proposes potential directions for future research. 
Our current focus is on time-series classification, where ROCKET stands as a state-of-the-art method. 
Given that \textsc{RockNet} is directly built upon ROCKET, its applicability however remains confined to this specific problem domain.

While time-series classification holds significant importance for CPS~\cite{fei2019cps,mohammadi2018deep,liang2019machine}, tasks such as image classification also represent significant research directions, to which ROCKET and consequently \textsc{RockNet} are not applicable directly.
To overcome this limitation, our approach could be adapted to fine-tune the final layer of pre-trained NN models.
As a last layer functions as a linear classifier on features extracted from earlier layers, the entire training procedure of \textsc{RockNet} (Steps 3--6 in Figure~\ref{fig:distributedrocketoverview}) can be utilized.
In this scenario, devices would share the computational load of the earlier layers, akin to how \textsc{RockNet} distributes the feature extraction process in ROCKET (Steps 1 and 2 in Figure~\ref{fig:distributedrocketoverview}). 
For convolutional NNs, this can be accomplished by leveraging existing methods for distributed inference~\cite{mao2017modnn, mao2017mednn, sahu2021denni}.

It is important to recognize that fine-tuning only the last layer of NNs results in lower accuracy compared to fine-tuning earlier layers, as demonstrated by Lin et al.~\cite{lin2022device}.
Our current approach does not support this level of training.
Consequently, exploring how multiple cooperating devices can fine-tune hidden layers of NNs, a process that would require communication during the backward pass, represents an interesting direction for future research. 
While our Mixer-based communication approach could serve as a foundation, substantial investigation is necessary to reduce communication overhead during the backward pass without compromising training performance.

At present, our method is backed up by the reported empirical evaluations. 
Exploring the theoretical properties of ROCKET, particularly in conjunction with the effects of quantization, offers an interesting direction for future research.

\section{Conclusions}

In this work, we have demonstrated that designing split learning at the intersection of ML and communication enables high accuracy training on ultra-low-power hardware.
The result of our design--\textsc{RockNet}--is the first TinyML split learning method capable of training non-linear timeseries classifiers on ultra-low-power hardware.
It integrates ROCKET classifiers with distributed compute and Mixer, a wireless communication protocol based on synchronous transmissions and network coding.
By efficiently pooling the resources of all devices within a CPS, this enables training of ROCKET classifiers, which are too large for a single ultra-low-power device.

Our experiments demonstrate that \textsc{RockNet} features a significantly higher accuracy than on-device central NN training.
\textsc{RockNet} scales well with number of devices, significantly reducing per-device memory consumption and latency.

Additionally, we demonstrated that ROCKET can be trained on ultra-low-power hardware. 
Although ROCKET delivers superior accuracy and lower resource consumption compared to classifiers such as HIVE-COTE, TS-CHIEF, and ResNet~\cite{dempster2020rocket,dempster2023hydra,dempster2021minirocket,tan2022multirocket}, its suitability for ultra-low-power contexts was previously unproven. 
Our parallelization strategy, merging ROCKET with Mixer, enables this and opens new avenues for deploying advanced ML algorithms on resource-constrained hardware.

In this work, we have not investigated the fault tolerance of our approach, e.g., regarding message loss and node failures. While the first seldom occurs due to Mixer's high reliability~\cite{Mixer}, the latter should be investigated in future work.

A promising direction for future research is to examine the interplay between \textsc{RockNet}'s online training capabilities and the efficiency gains achievable through ROCKET pruning methods \cite{giordano2023optimizing}. Combining these techniques post-training may further reduce resource consumption during inference.

Finally, our split learning approach is not limited to ROCKET classifiers. Mixer is generally well-suited for distributed AI in low-power, multi-hop networks, suggesting that future research should explore how other ML algorithms can be effectively distributed across CPS devices using this protocol.

\begin{acks}
We thank Andrés Posada Moreno and Friedrich Solowjow for insightful discussions, and Alperen Cebecik for his support with implementing the approach.

This work was supported by the German Research Foundation (DFG) within the priority program 1914 (grant TR 1433/2) and within the Emmy Noether project NextIoT (ZI 1635/2-1), and by the LOEWE initiative (Hesse, Germany) within the emergenCITY center (LOEWE/1/12/519/03/05.001(0016)/72). 

The authors gratefully acknowledge the computing time provided to them at the NHR Center NHR4CES at RWTH Aachen University (p0021919). This is funded by the Federal Ministry of Education and Research, and the state governments participating on the basis of the resolutions of the GWK for national high performance computing at universities (www.nhr-verein.de/unsere-partner).

\end{acks}

\bibliographystyle{ACM-Reference-Format}
\bibliography{references}

\clearpage
\appendix
\appendix

\section{I.I.D vs Non-I.I.D Distributed Datasets}
\label{app:iid}

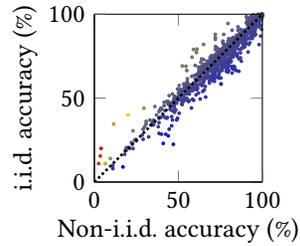
\begin{figure}[h]
    \centering
    \begin{subfigure}[b]{0.48\linewidth}
      \centering
      \begin{tikzpicture}[spy using outlines={circle, magnification=3, size=2cm, connect spies, every spy on node/.append style={thick}}]
      \begin{axis}[
        xmax=100.01, xmin=-0.01,
        ymax=100.01, ymin=-0.01,
        name=plot3, xlabel={Non-i.i.d. accuracy (\SI{}{\percent})}, 
        ylabel={i.i.d. accuracy (\SI{}{\percent})},
        ylabel style={align=center,yshift=-1mm},
        xlabel style={align=center,yshift=1mm, xshift=0mm},
        height=0.5\linewidth, width=0.5\linewidth,
        clip mode=individual,
        view={0}{90},
        ]

        \addplot3[scatter,mark=*, mark size=0.5pt, only marks] table [x=accNonIID, y=accIID, z=distanceBoundary, col sep=comma] {images/ComparisonIID.csv};
  
        \addplot[color=black, line width=1pt, mark=none, dotted, dash pattern=on 1pt off 1pt] coordinates {(0,0) (100,100)};
      \end{axis}
    \end{tikzpicture}
  \end{subfigure}
  \caption{Ablation i.i.d. vs. non-i.i.d distributed datasets. The experiment demonstrates that \textsc{RockNet} is robust against non-i.i.d. data.}
  \label{fig:iid}
  \end{figure}

We conducted experiment on the UCR dataset with i.i.d. sampled data and non-i.i.d. sampled data.
For the non-i.i.d. data, we sorted the dataset according to its indexes. Then, we split it evenly among 10 devices. Through this, a device only holds datapoints form a few classes and different devices hold data from different classes.
Our results shown in Figure~\ref{fig:iid} demonstrate that \textsc{RockNet} retains the same accuracy, when trained on non-i.i.d. sampled data, compared to i.i.d. sampled data.

\end{document}